\documentclass[letterpaper, 10 pt, journal, twoside]{IEEEtran}
\IEEEoverridecommandlockouts
\usepackage{fancyhdr}
\usepackage[space]{cite}
\usepackage{amsmath,amsopn,amstext,amsfonts,mathtools}
\usepackage{amssymb}
\usepackage{float}
\usepackage{siunitx}
\usepackage{graphicx}
\usepackage{textcomp}
\usepackage{xcolor}
 \usepackage{multirow}
\usepackage{booktabs}
\usepackage{svg}
\usepackage[tight]{subfigure}
\usepackage{epstopdf}
\usepackage[linesnumbered,boxed,ruled,commentsnumbered]{algorithm2e}
\usepackage{algpseudocode}
\usepackage[font=small,skip=5pt]{caption}
\usepackage{geometry}
\def\BibTeX{{\rm B\kern-.05em{\sc i\kern-.025em b}\kern-.08em
    T\kern-.1667em\lower.7ex\hbox{E}\kern-.125emX}}

\begin{document}

\title{Targetless Extrinsic Calibration of Multiple Small FoV LiDARs and Cameras using Adaptive Voxelization}

\author{Xiyuan Liu, Chongjian Yuan, and Fu Zhang
\thanks{Xiyuan Liu, Chongjian Yuan and Fu Zhang are with the Department of Mechanical Engineering, The University of Hong Kong, Pokfulam, Hong Kong Special Administrative Region, People's Republic of China. (Corresponding author: Fu Zhang) (email: {\tt\footnotesize $\{$xliuaa,ycj1$\}$@connect.hku.hk}, {\tt\footnotesize $ $fuzhang$ $@hku.hk}).}
}
\maketitle
\thispagestyle{empty}
\begin{abstract}
Determining the extrinsic parameter between multiple LiDARs and cameras is essential for autonomous robots, especially for solid-state LiDARs, where each LiDAR unit has a very small Field-of-View (FoV), and multiple units are often used collectively. The majority of extrinsic calibration methods are proposed for 360$^\circ$ mechanical spinning LiDARs where the FoV overlap with other LiDAR or camera sensors is assumed. Few research works have been focused on the calibration of small FoV LiDARs and cameras nor on the improvement of the calibration speed. In this work, we consider the problem of extrinsic calibration among small FoV LiDARs and cameras, with the aim to shorten the total calibration time and further improve the calibration precision. We first implement an adaptive voxelization technique in the extraction and matching of LiDAR feature points. Such a process could avoid the redundant creation of $k$-d trees in LiDAR extrinsic calibration and extract LiDAR feature points in a more reliable and fast manner than existing methods. We then formulate the multiple LiDAR extrinsic calibration into a LiDAR Bundle Adjustment (BA) problem. By deriving the cost function up to second-order, the solving time and precision of the non-linear least square problem are further boosted. Our proposed method has been verified on data collected in four targetless scenes and under two types of solid-state LiDARs with a completely different scanning pattern, density, and FoV. The robustness of our work has also been validated under eight initial setups, with each setup containing 100 independent trials. Compared with the state-of-the-art methods, our work has increased the calibration speed 15 times for LiDAR-LiDAR extrinsic calibration (averaged result from 100 independent trials) and 1.5 times for LiDAR-Camera extrinsic calibration (averaged result from 50 independent trials) while remaining accurate. To benefit the robotics community, we have also open-sourced our implementation code on GitHub.
\end{abstract}

\begin{IEEEkeywords}
	Multiple LiDAR-Camera Extrinsic Calibration, Small FoV LiDAR, High-Resolution Mapping.
\end{IEEEkeywords}

\section{Introduction}
LiDAR and camera sensors, due to their superior characteristics in direct spatial ranging and rich color information conveying, have been increasingly used in autonomous driving~\cite{avoid2021kong,lin2020de}, navigation~\cite{loamlivox,balm} and high-resolution mapping~\cite{previous_work} applications. One drawback of the current 360$^\circ$ mechanical spinning LiDAR is their dramatic high cost, preventing their massive application in industry. Solid-state LiDAR~\cite{lowcost} has a much lower cost while achieving a denser point cloud within its FoV. However, solid-state LiDARs are of small FoV that multiple solid-state LiDARs need to be combined to achieve a similar FoV coverage as the mechanical spinning LiDAR. This setup necessitates precise extrinsic calibration among the LiDARs and cameras.

\begin{figure}[t]
    \centering
    \includegraphics[width=1.0\linewidth]{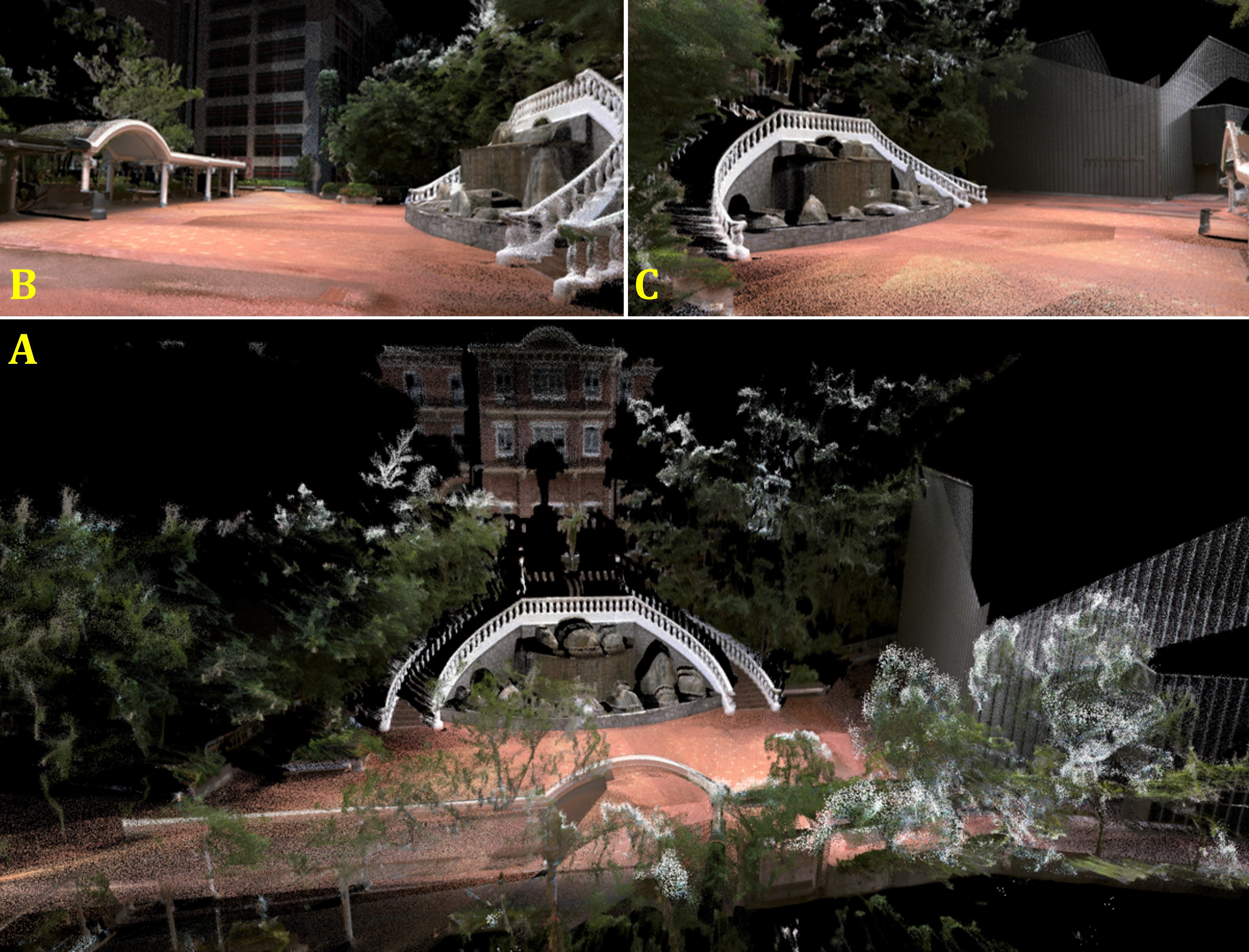}
    \caption{A) The dense colorized point cloud with the LiDAR poses and extrinsic parameters optimized by our proposed method. The views from other perspectives are exhibited in B) left side and C) right side. Our experiment video is available at https://youtu.be/PaiYgAXl9iY.}
    \label{fig:cover_photo}
\end{figure}

Several challenges reside in the extrinsic calibration involving small FoV LiDARs: (1) Limited FoV overlap among the sensors and the precision requirement. Current methods usually require the existence of a common FoV between each pair of sensors~\cite{online2010gao,automatic2019xue,levinson2013automatic,jfr_mutual_info}, such that each feature is viewed by all sensors. In real-world applications, this FoV overlap might be minimal or not even exist due to the small FoVs of solid-state LiDARs and their numerous sensor mounting positions. The accuracy requirement of the calibration results, e.g., the consistency and colorization of the point cloud (see Fig.~\ref{fig:cover_photo}), is thus more challenging. (2) Computation time demands. For general ICP-based LiDAR extrinsic calibration approaches~\cite{robust2021jiao,previous_work}, the extrinsic is optimized by aligning the point cloud from all LiDARs and maximizing the point cloud’s consistency. The increase in the number of LiDARs implies that the feature point correspondence searching will be more time-consuming. This is due to the reason that each feature point needs to search for and match with nearby feature points using a $k$-d tree which contains the whole point cloud. In the LiDAR-camera extrinsic calibration, a larger amount of LiDAR points will also lead to more computation time in the LiDAR feature extraction.

To address the above challenges, we propose a fast and targetless approach for extrinsic calibration of multiple small FoV LiDARs and cameras. To create enough co-visible features among the small FoV sensors, we introduce motions to the sensor platform such that each sensor will scan the same area (hence features) at different times. We first calibrate the extrinsic among LiDARs (and simultaneously estimate the LiDAR poses) by registering their point cloud using an efficient Bundle Adjustment (BA) method we recently proposed~\cite{balm}. To reduce time consumption in feature correspondence matching among LiDARs, we implement an adaptive voxelization to dynamically segment the point cloud into multiple voxels so that only one plane feature resides in each voxel (see Sec.~\ref{sec:voxel}). We then calibrate the extrinsic between the cameras and LiDARs by matching the co-visible features between the images and the above-reconstructed point cloud. To further accelerate the feature correspondence matching, we inherit the above adaptive voxel map to extract LiDAR edge features. In summary, our contributions are listed as follows:

\begin{itemize}
\item We propose a targetless extrinsic calibration pipeline for multiple small FoV LiDARs and cameras that share very few or even no FoV overlap. We formulate LiDAR extrinsic calibration into a Bundle Adjustment problem and implement an adaptive voxelization technique into the LiDAR feature extraction and matching process. The overall pipeline enjoys higher calibration precision and computation efficiency.
\item We verify our proposed work on data collected in various test scenes by LiDARs of different scanning patterns, FoVs, and point densities. When compared to various state-of-the-art methods, our proposed work could boost the speed by 15 times for multiple LiDAR calibration and 1.5 times for multiple LiDAR-Camera calibration. Meanwhile, our proposed work maintains high calibration precision, with the average translation and rotation errors down to 6mm and 0.09 degrees for LiDAR-camera and 8mm and 0.2 degrees for LiDAR-LiDAR.
\item We open-source our implementation in ROS on GitHub\footnote{https://github.com/hku-mars/mlcc} to benefit the robotics community.
\end{itemize}

\section{Related Works}\label{sec:related_works}
\subsection{LiDAR-LiDAR Extrinsic Calibration}
The extrinsic calibration methods between multiple LiDARs could be divided into motion-based and motionless approaches. Motion-based approaches assume each sensor undergoes the same rigid motion in each time interval~\cite{automatic2020heng,tro_motion,lin2020de} and transform the extrinsic calibration into a Hand-Eye problem~\cite{handeye}. Authors in~\cite{levinson2014,lost2012,billah2019} also introduce external inertial navigation sensors to facilitate the motion estimation of LiDARs. The calibration precision of these approaches is easily affected by the accuracy of the LiDAR odometry results, which might be unreliable. Motionless methods have been discussed in~\cite{online2010gao,automatic2019xue} where the authors attach retro-reflective tapes to the surface of calibration targets to create and facilitate the feature extraction among multiple LiDARs. These approaches require prior preparation work and FoV overlap between LiDARs, which is unpractical in real-world applications.

In our previous work~\cite{previous_work}, a simple rotational movement is introduced to eliminate the requirement of FoV overlap, as each onboard sensor could percept the same region of interest. Then the extrinsic parameter is calibrated, along with the estimation of LiDAR poses, by optimizing the consistency of the point cloud map with iterative closest point (ICP) registration. The main problem within~\cite{previous_work} is that the ICP registration always registers one scan to the other, leading to an iterative process where only one optimization variable (e.g., extrinsic or LiDAR poses) can be optimized (by registering the point cloud affected by the variable under optimization to the rest). Such an iterative procedure is prolonged to converge. Moreover, at each iteration, the ICP-based feature correspondence matching process might be very time-consuming. As for each point-to-plane correspondence, ICP needs to either search inside a $k$-d tree containing the entire point cloud or create a $k$-d tree containing the local point cloud every time before searching.

In this work, we formulate the extrinsic calibration into a bundle adjustment (BA) problem~\cite{balm}, where all the optimization variables (both extrinsic and LiDAR poses) are optimized concurrently by registering points into their corresponding plane. When compared to other plane adjustment techniques~\cite{planeAdjust,lips}, the BA technique we use does not estimate the plane parameters in the optimization process but solves for them analytically in a closed-form solution prior to the optimization iteration. The removal of plane parameters from the optimization iteration lowers the dimension significantly and leads to very efficient multi-view registration. To match points corresponding to the same plane, we implement an adaptive voxelization technique \cite{balm} to replace the $k$-d tree in \cite{previous_work}. As only one plane feature exists in each voxel, our proposed work significantly saves the computation time in correspondence searching while remaining accurate (see Sec.~\ref{sec:voxel}).

\subsection{LiDAR-Camera Extrinsic Calibration}
The extrinsic calibration between LiDAR and camera could be mainly divided into target-based and targetless methods. In target-based approaches, the geometric features, e.g., edges and surfaces, are extracted from artificial geometric solids~\cite{kummerle2020,Rodriguez2008,park2014} or chessboard~\cite{koo2020analytic,zhou2018automatic} using intensity and color information. These features are matched either automatically or manually and are solved with non-linear optimization tools. In~\cite{roadIsEnough}, authors establish the constraints using the crosswalk features on the streets; however, this method is essentially target-based as the parallelism characteristic of the crosswalk is used. Since extra calibration targets and manual work are needed, these methods are less practical compared with targetless solutions.

The targetless methods could be further divided into motion-based and motionless approaches. In motion-based methods, the initial extrinsic parameter is usually estimated by the motion information and refined by the appearance information. In~\cite{nagy2019}, authors reconstruct a point cloud from images using the structure from motion (SfM) to determine the initial extrinsic parameter and refine it by back-projecting LiDAR points onto the image plane. In~\cite{tro_motion,spatiotemporal}, authors initialize the extrinsic parameter by Hand-Eye calibration and optimize it by minimizing the re-projection error between images and LiDAR scans. In motionless approaches, only the edge features that co-exist in both sensors' FoV are extracted and matched. Then the extrinsic parameter is optimized by minimizing the re-projected edge-to-edge distances~\cite{pixel_level,camvox,scaramuzza2007extrinsic,levinson2013automatic} or by maximizing the mutual information between the back-projected LiDAR points and the images~\cite{jfr_mutual_info}.

Our proposed work is targetless and creates co-visible features by moving the sensor suite to multiple poses, hence allowing extrinsic calibration between LiDAR and cameras even when they have no overlap, a circumstance that was not solved in prior works~\cite{tro_motion,jfr_mutual_info,camvox}. Moreover, compared with our previous work~\cite{pixel_level} which extracts LiDAR edge features using the RANSAC algorithm, this work extracts edge features using the same adaptive voxelization already computed in the LiDAR extrinsic calibration, which is more competitive in computation time and calibration precision. Compared with~\cite{jfr_mutual_info} which uses LiDAR intensity information as a feature, our work uses more reliable 3D edge information and is more computationally efficient and accurate (see Sec.~\ref{sec:experiments}). Moreover, our work does not require the common FoV between sensors.

\section{Methodology}\label{sec:methodology}
\subsection{Overview}\label{sec:overview}
\begin{figure}[t]
    \centering
    \includegraphics[width=0.7\linewidth]{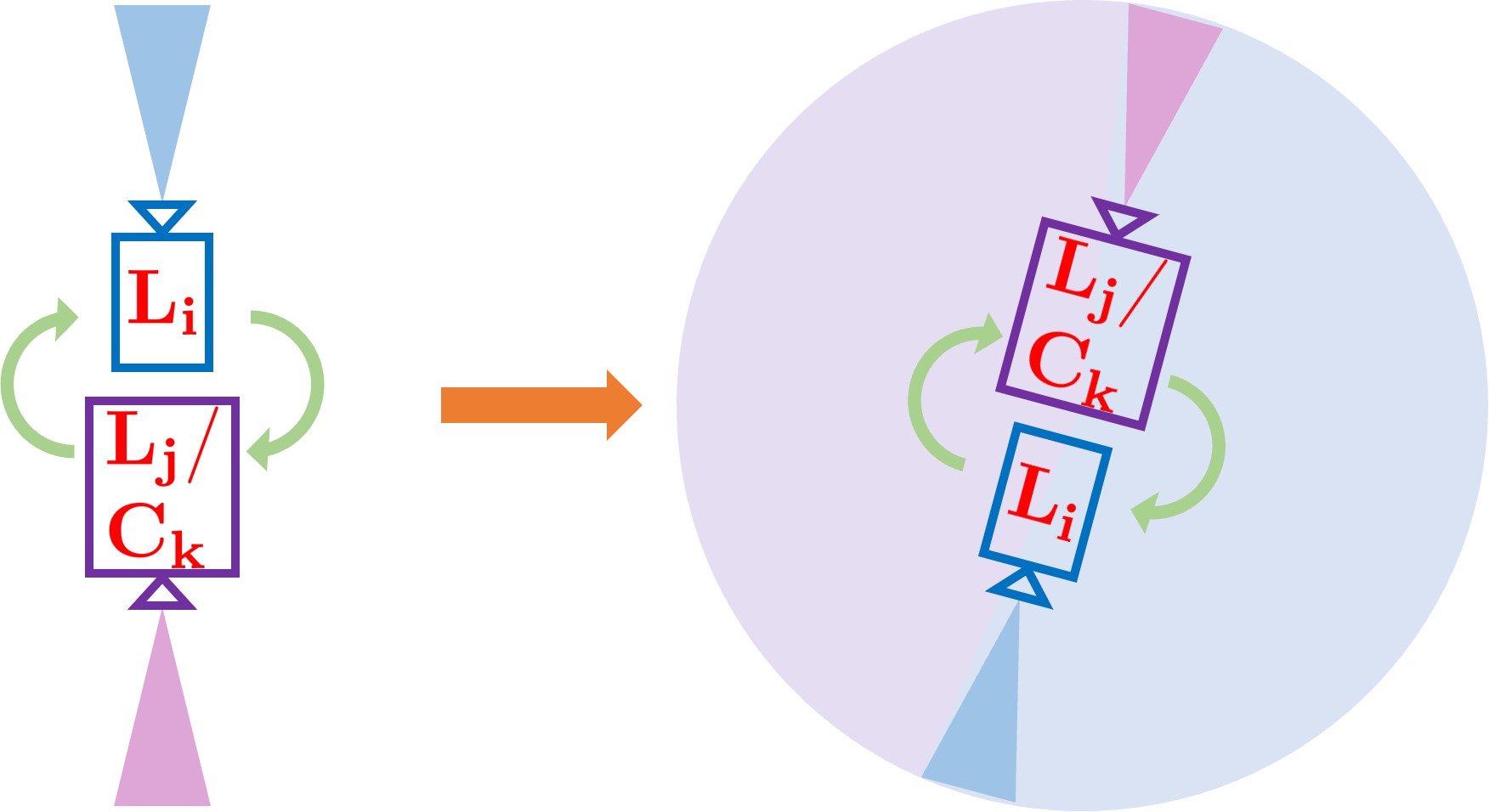}
    \caption{FoV overlap created by rotation between two opposite pointing sensors. The original setup of two sensors $L_i$ and $L_j/C_k$ share no FoV overlap. With the introduction of rotational motion, the same region is scanned by all sensors across different times.}
    \label{fig:rotate_fov}
\end{figure}

Let $\prescript{B}{A}{\mathbf{T}}=(\prescript{B}{A}{\mathbf{R}},\prescript{B}{A}{\mathbf{t}})\in SE(3)$ represent the rigid transformation from frame $A$ to frame $B$, where $\prescript{B}{A}{\mathbf{R}}\in SO(3)$ and $\prescript{B}{A}{\mathbf{t}}\in\mathbb{R}^3$ are the rotation and translation. We denote $\mathcal{L}=\{L_0,L_1,\cdots,L_{n-1}\}$ the set of $n$ LiDARs, where $L_0$ represents the base LiDAR for reference, $\mathcal{C}=\{C_0,C_1,\cdots,C_h\}$ the set of $h$ cameras, $\mathcal{E}_L=\{\prescript{L_0}{L_1}{\mathbf{T}},\prescript{L_0}{L_2}{\mathbf{T}},\cdots,\prescript{L_0}{L_{n-1}}{\mathbf{T}}\}$ the set of LiDAR extrinsic parameters and $\mathcal{E}_C=\{\prescript{C_0}{L_0}{\mathbf{T}},\prescript{C_1}{L_0}{\mathbf{T}},\cdots,\prescript{C_h}{L_0}{\mathbf{T}}\}$ the set of LiDAR-camera extrinsic parameters. To create co-visible features between multiple LiDARs and cameras that may share no FoV overlap, we rotate the robot platform to $m$ poses such that the same region of interest is scanned by all sensors (see Fig.~\ref{fig:rotate_fov}). Denote $\mathcal{T}=\{t_0,t_1,\cdots,t_{m-1}\}$ the time for each of the $m$ poses and the pose of the base LiDAR at the initial time as the global frame, i.e., $\prescript{G}{L_0}{\mathbf{T}}_{t_0}=\mathbf{I}_{4\times4}$. Denote $\mathcal{S}=\{\prescript{G}{L_0}{\mathbf{T}}_{t_1},\prescript{G}{L_0}{\mathbf{T}}_{t_2},\cdots,\prescript{G}{L_0}{\mathbf{T}}_{t_{m-1}}\}$ the set of the base LiDAR poses in global frame. The point cloud patch scanned by LiDAR $L_i\in\mathcal{L}$ at time $t_j\in\mathcal{T}$ is denoted by $\mathcal{P}_{L_i,t_j}$, which is in $L_i$'s local frame. This point cloud patch could be transformed to global frame by
\begin{equation}
    \begin{split}
    \prescript{G}{}{\mathcal{P}}_{L_i,t_j} &= \prescript{G}{L_i}{\mathbf{T}}_{t_j} \mathcal{P}_{L_i,t_j} \\
    &\triangleq \{ \prescript{G}{L_i}{\mathbf{R}}_{t_j} \mathbf{p}_{L_i,t_j} + \prescript{G}{L_i}{\mathbf{t}}_{t_j}, \ \forall \mathbf{p}_{L_i,t_j} \in \mathcal{P}_{L_i,t_j}  \}.
    \end{split}
\end{equation}

In our proposed approach of multi-sensor calibration, we sequentially calibrate the $\mathcal{E}_L$ and $\mathcal{E}_C$. In the first step, we simultaneously estimate the LiDAR extrinsic $\mathcal{E}_L$ and the base lidar pose trajectory $\mathcal{S}$ based on an efficient multi-view registration (see Sec.~\ref{sec:methodology_multi_lidar}). In the second step, we calibrate the $\mathcal{E}_C$ by matching the depth-continuous edges extracted from images and the above-reconstructed point cloud (see Sec.~\ref{sec:methodology_lidar_camera}). Lying in the center of both LiDAR and camera extrinsic calibration is an adaptive map, which finds correspondence among LiDAR and camera measurements efficiently (Sec. \ref{sec:voxel}).

\subsection{Adaptive Voxelization} \label{sec:voxel}
To find the correspondences among different LiDAR scans, we assume the initial base LiDAR trajectory $\mathcal{S}$, LiDAR extrinsic $\mathcal{E}_L$, and camera extrinsic $\mathcal{E}_C$ are available. The initial base LiDAR trajectory $\mathcal{S}$ could be obtained by an online LiDAR SLAM (e.g., \cite{loamlivox}), and the initial extrinsic could be obtained from the CAD design or a rough Hand-Eye calibration~\cite{handeye}. Our previous work \cite{previous_work} extracts edge and plane feature points from each LiDAR scan and matches them to the nearby edge and plane points in the map by a $k$-nearest neighbor search ($k$-NN). This would repeatedly build a $k$-d tree of the global map at each iteration. In this paper, we use a more efficient voxel map proposed in \cite{balm} to create correspondences among all LiDAR scans.

The voxel map is built by cutting the point cloud (registered using the current $\mathcal{S}$ and $\mathcal{E}_L$) into small voxels such that all points in a voxel roughly lie on a plane (with some adjustable tolerance). The main problem of the fixed-resolution voxel map is that if the resolution is high, the segmentation would be too time-consuming, while if the resolution is too low, multiple small planes in the environments falling into the same voxel would not be segmented. To best adapt to the environment, we implement an adaptive voxelization process. More specifically, the entire map is first cut into voxels with a pre-set size (usually large, e.g., 4m). Then for each voxel, if the contained points from all LiDAR scans roughly form a plane (by checking the ratio between eigenvalues), it is treated as a planar voxel; otherwise, they will be divided into eight octants, where each will be examined again until the contained points roughly form a plane or the voxel size reaches the pre-set minimum lower bound. Moreover, the adaptive voxelization is performed directly on the LiDAR raw points, so no prior feature points extraction is needed as in \cite{previous_work}.

\begin{figure}[t]
    \centering
    \includegraphics[width=1.0\linewidth]{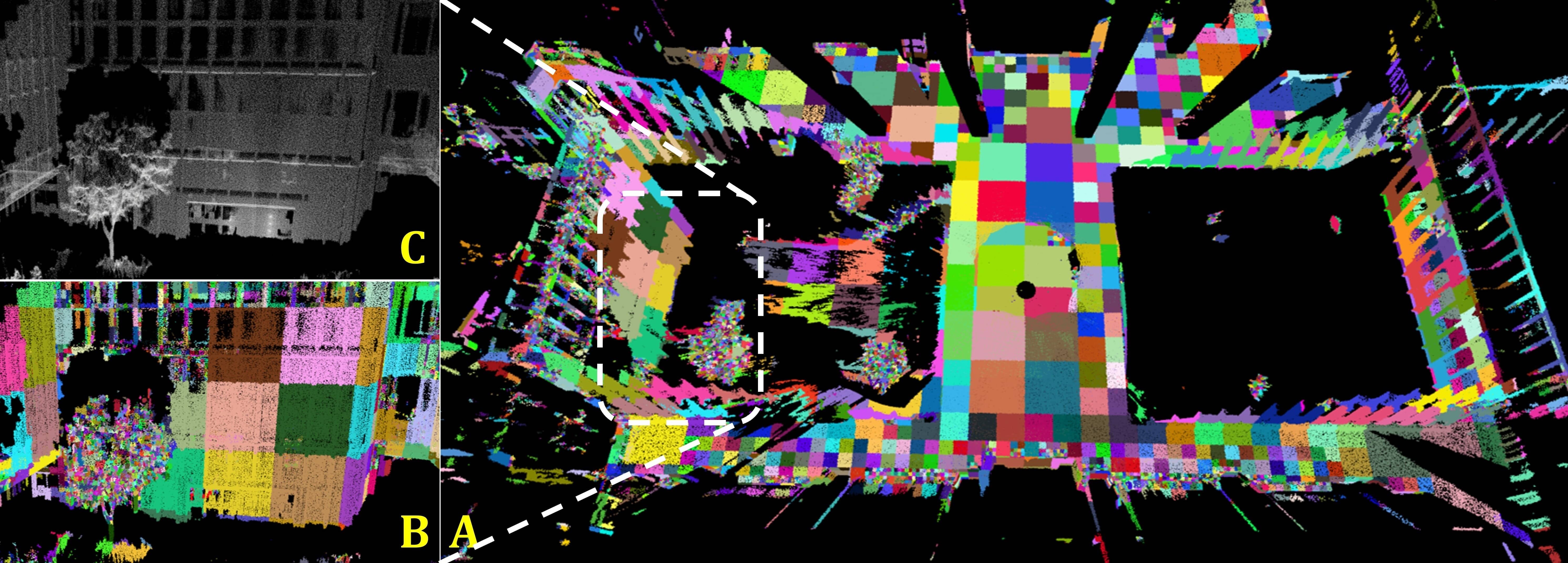}
    \caption{A) LiDAR point cloud segmented with the adaptive voxelization. Points within the same voxel are colored identically. The detailed adaptive voxelization of points in the dashed white rectangle could be viewed in B) colored points and C) original points. The default size for the initial voxelization is 4m, and the minimum voxel size is 0.25m.}
    \label{fig:voxel_lidar}
\end{figure}

Fig.~\ref{fig:voxel_lidar} shows a typical result of the adaptive voxelization process in a complicated campus environment. As can be seen, this process is able to segment planes of different sizes, including large planes on the ground, medium planes on the building walls, and tiny planes on tree crowns.

\subsection{Multi-LiDAR Extrinsic Calibration}\label{sec:methodology_multi_lidar}

\begin{figure}[t]{
\centering
\subfigure[]{
\begin{minipage}[t]{0.48\linewidth}
\centering
\includegraphics[width=1\linewidth]{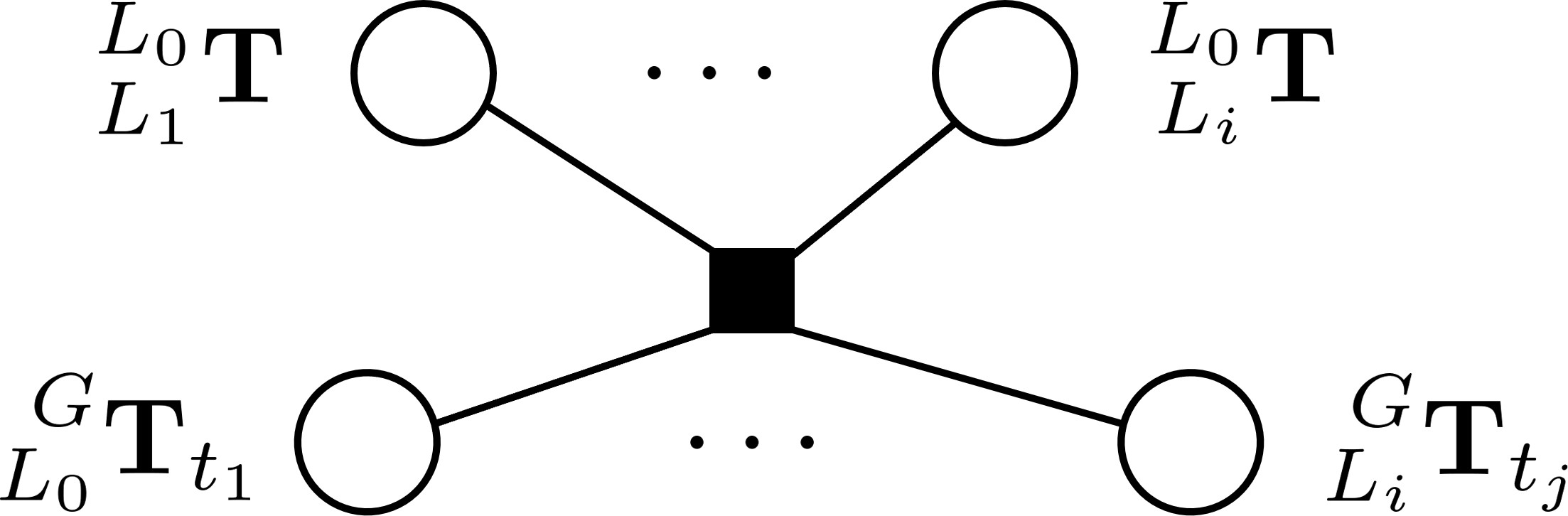}
\vspace{-0.5cm}
\label{fig:factor}
\end{minipage}
}%
\subfigure[]{
\begin{minipage}[t]{0.48\linewidth}
\centering
\includegraphics[width=1\linewidth]{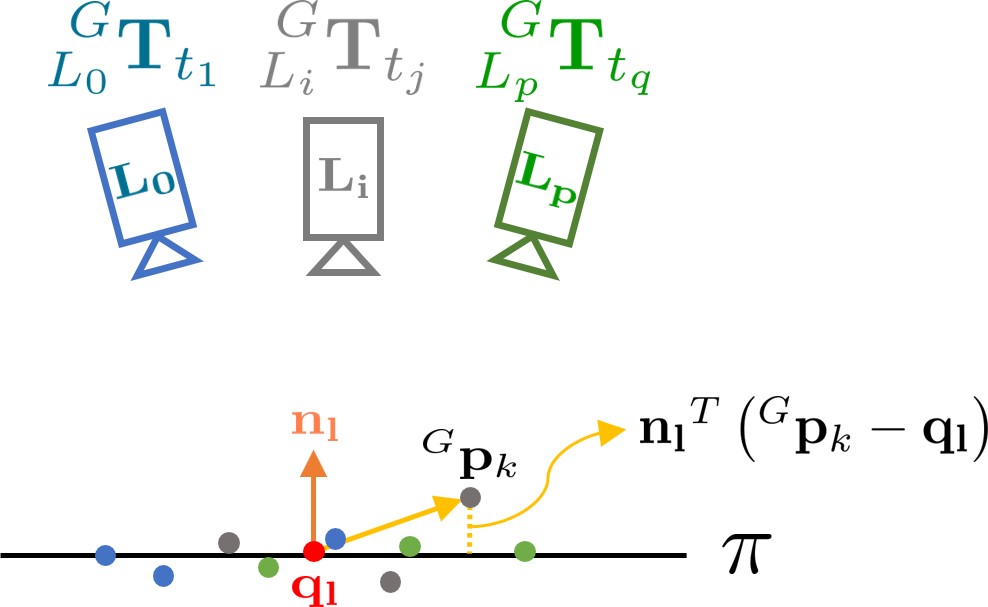}
\vspace{-0.5cm}
\label{fig:point2plane}
\end{minipage}%
}%
}
\centering
\caption{(a) The $l$-th factor item relating to $\mathcal{S}$ and $\mathcal{E}_L$ with $L_i\in\mathcal{L}$ and $t_j\in\mathcal{T}$. (b) The distance from the point $\prescript{G}{}{\mathbf{p}}_k$ to the plane $\boldsymbol{\pi}$.}
\label{fig:l_factor}
\end{figure}

With adaptive voxelization, we can obtain a set of voxels of different sizes. Each voxel contains points that are roughly on a plane and creates a planar constraint for all LiDAR poses that have points in this voxel. More specifically, considering the $l$-th voxel consisting of a group of points $\mathcal{P}_l=\{\prescript{G}{}{\mathbf{p}}_{L_i,t_j}\}$ scanned by $L_i \in \mathcal{L}$ at times $t_j \in \mathcal{T}$. We define a point cloud consistency indicator $c_l\left(\prescript{G}{L_i}{\mathbf{T}}_{t_j}\right)$ which forms a factor on $\mathcal{S}$ and $\mathcal{E}_L$ as shown in Fig.~\ref{fig:factor}. Then, the base LiDAR trajectory and extrinsic are estimated by optimizing the factor graph. A natural choice for the consistency indicator $c_l\left(\cdot\right)$ would be the summed Euclidean distance between each $\prescript{G}{}{\mathbf{p}}_{L_i,t_j}$ to the plane to be estimated (see Fig.~\ref{fig:point2plane}). Taking account of all such indicators within the voxel map, we could formulate the problem as
\begin{equation}\label{eq:l_factor}
    \arg\min_{\mathcal{S},\mathcal{E}_{L}, \mathbf n_l, \mathbf q_l}\sum_l \underbrace{\left(\frac{1}{N_l}\sum_{k=1}^{N_l}\left(\mathbf{n}_l^T\left(\prescript{G}{}{\mathbf{p}}_k-\mathbf{q}_l\right)\right)^2\right)}_{l\text{-th factor}},
\end{equation}
where $\prescript{G}{}{\mathbf{p}}_k\in \mathcal{P}_l$, $N_l$ is the total number of points in $\mathcal{P}_l$, $\mathbf{n}_l$ is the normal vector of the plane and $\mathbf{q}_l$ is a point on this plane.

It is noticed that the optimization variables $(\mathbf n_l, \mathbf q_l)$ in~\eqref{eq:l_factor} could be analytically solved (see Appendix A) and the resultant cost function~\eqref{eq:lambda_3_A} is over the LiDAR pose $\prescript{G}{L_i}{\mathbf{T}}_{t_j}$ (hence the base LiDAR trajectory $\mathcal{S}$ and extrinsic $\mathcal{E}_L$) only, as follows
\begin{equation}\label{eq:lambda_3_A}
    \arg\min_{\mathcal{S},\mathcal{E}_{L}}\sum_l \lambda_3\left(\mathbf A_l\right)
\end{equation}
where $\lambda_3(\mathbf A_l)$ denotes the minimal eigenvalue of matrix $\mathbf A_l$ defined as
\begin{equation}
\mathbf{A}_l=\frac{1}{N_l}\sum_{k=1}^{N_l}\prescript{G}{}{\mathbf{p}}_k\prescript{G}{}{\mathbf{p}}_k^T-\mathbf{q}_l^{\ast} \mathbf{q}_l^{\ast T},\mathbf{q}_l^{\ast}=\frac{1}{N_l}\sum_{k=1}^{N_l}\prescript{G}{}{\mathbf{p}}_k.
\end{equation}
To allow efficient optimization in~\eqref{eq:lambda_3_A}, we derive the closed-form derivatives w.r.t the optimization variable $\mathbf{x}$ up to second-order (the detailed derivation from~\eqref{eq:lambda_3_A} to~\eqref{eq:lamdba_3_x} is elaborated in Appendix B):
\begin{equation}\label{eq:lamdba_3_x}
    \lambda_3(\mathbf{x}\boxplus\mathbf{\delta x})\approx
    \lambda_3(\mathbf{x})+\mathbf{\Bar{J}}\delta\mathbf{x}+\frac{1}{2}\delta\mathbf{x}^T\mathbf{\Bar{H}}\delta\mathbf{x},
\end{equation}
where $\Bar{\mathbf{J}}$ is the Jacobian matrix, and $\Bar{\mathbf{H}}$ is the Hessian matrix. The $\delta{\mathbf{x}}$ is a small perturbation of the optimization variable $\mathbf x$:
\begin{equation*}
\begin{aligned}
    \mathbf{x}&=[\underbrace{\cdots\prescript{G}{L_0}{\mathbf{R}}_{t_j}\ \prescript{G}{L_0}{\mathbf{t}}_{t_j}\cdots}_{\mathcal{S}}\underbrace{\cdots\prescript{L_0}{L_i}{\mathbf{R}}\ \prescript{L_0}{L_i}{\mathbf{t}}\cdots}_{\mathcal{E}_L}]. \\
\end{aligned}
\end{equation*}
Then the optimal $\mathbf{x}^*$ could be determined by iteratively solving~\eqref{eq:HxJ} with the LM method and updating the $\delta\mathbf{x}$ to $\mathbf{x}$.
\begin{equation}\label{eq:HxJ}
    \left(\Bar{\mathbf{H}}+\mu\mathbf{I}\right)\delta\mathbf{x}=-\Bar{\mathbf{J}}^T
\end{equation}

\subsection{LiDAR-Camera Extrinsic Calibration}\label{sec:methodology_lidar_camera}
With the LiDAR extrinsic parameter $\mathcal{E}_L$ and pose trajectory $\mathcal{S}$ computed above, we obtain a dense global point cloud by transforming all LiDAR points to the base LiDAR frame. Then, the extrinsic $\mathcal{E}_C$ is optimized by minimizing the summed distance between the back-projected LiDAR edge feature points and the image edge feature points. Two types of LiDAR edge points could be extracted from the point cloud. One is the depth-discontinuous edge between the foreground and background objects, and the other is the depth-continuous edge between two neighboring non-parallel planes. As explained in our previous work~\cite{pixel_level}, depth-discontinuous edges suffer from foreground inflation and bleeding points phenomenon; we hence use depth-continuous edges to match the point cloud and images.

\begin{figure}[ht]
    \centering
    \includegraphics[width=1.0\linewidth]{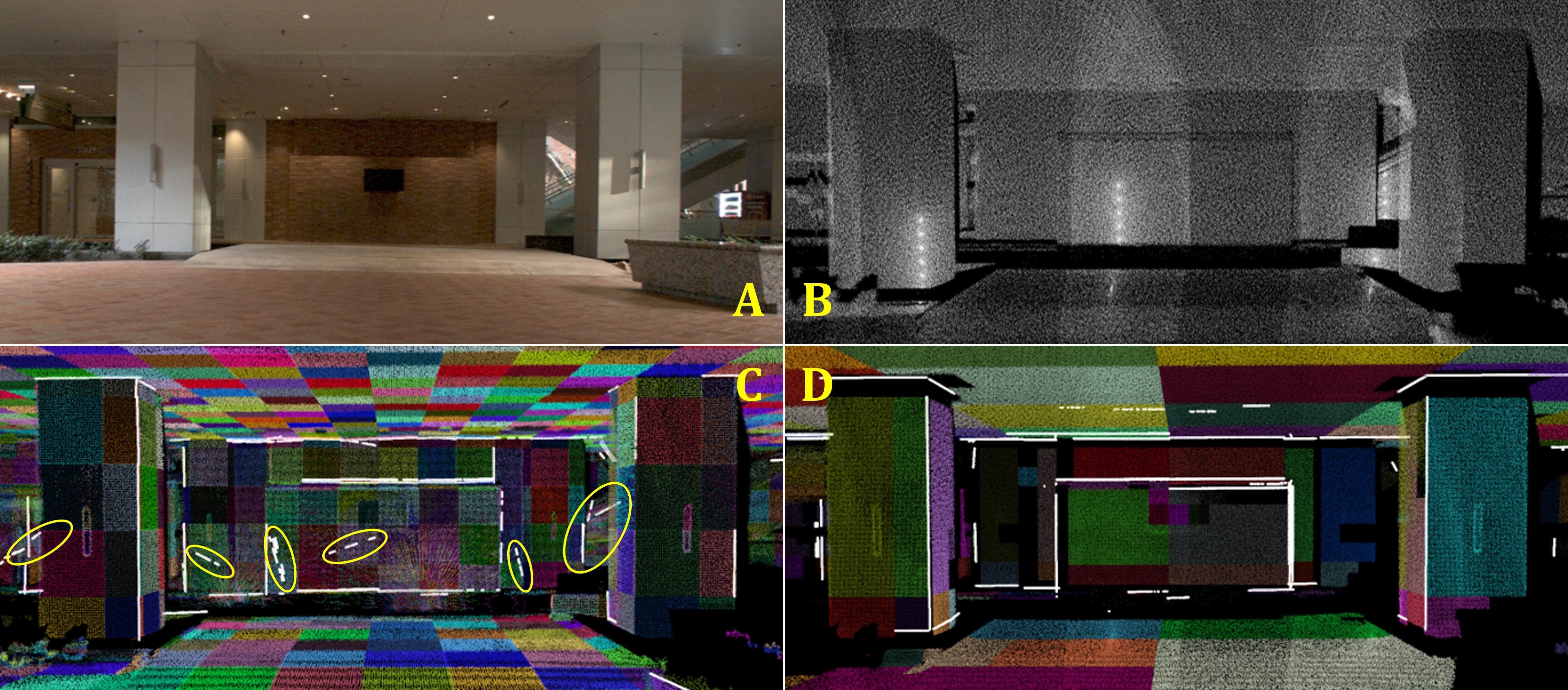}
    \caption{Depth-continuous LiDAR edge feature extraction comparison. A) Real-world image. B) Raw point cloud of this scene. C) Edges extracted using method in~\cite{pixel_level} where the yellow circles indicate the false estimations. D) Edges extracted with adaptive voxelization.}
    \label{fig:camera_voxel}
\end{figure}

In~\cite{pixel_level}, the LiDAR point cloud is segmented into voxels with uniform sizes, and the planes inside each voxel are estimated by the RANSAC algorithm. In contrast, our method uses the same adaptive voxel map obtained in Sec.~\ref{sec:voxel}. We calculate the angle between their containing plane normals for every two adjacent voxels. If this angle exceeds a threshold, the intersection line of these two planes is extracted as the depth-continuous edge, as shown in Fig.~\ref{fig:camera_voxel}. We choose to implement the Canny algorithm for image edge features to detect and extract.

Suppose $\prescript{G}{}{\mathbf{p}}_i$ represents the $i$-th point from a LiDAR edge feature extracted above in global frame. With pin-hole camera and its distortion model, $\prescript{G}{}{\mathbf{p}_i}$ is projected onto the image taken by camera $C_l$ at $t_j$, i.e., $\mathbf{I}_{l,j}$ by
\begin{equation}\label{eq:projection}
    \prescript{\mathbf{I}_{l,j}}{}{\mathbf{p}_i}=\mathbf{f}\left(\boldsymbol{\pi}\left(\prescript{C_l}{L_0}{\mathbf{T}}\left(\prescript{G}{L_0}{\mathbf{T}}_{t_j}\right)^{-1}\prescript{G}{}{\mathbf{p}}_i\right)\right),
\end{equation}

\noindent
where $\mathbf{f(\boldsymbol{\cdot})}$ is the camera distortion model and $\boldsymbol{\pi(\cdot)}$ is the projection model. Let $\mathcal{I}_i$ represent the set of images that capture the point $\prescript{G}{}{\mathbf{p}_i}$, i.e., $\mathcal{I}_i=\{\mathbf{I}_{l,j}\}$. For each $\prescript{\mathbf{I}_{l,j}}{}{\mathbf{p}_i}$, the $\kappa$ nearest image edge feature points $\mathbf{q}_k$ on $\mathbf{I}_{l,j}$ are searched. The normal vector $\mathbf{n}_{i,l,j}$ of the edge formed by these $\kappa$ points is thus the eigenvector corresponding to the minimum eigenvalue of $\mathbf{A}_{i,l,j}$ that
\begin{equation}
    \mathbf{A}_{i,l,j}=\sum^{\kappa}_{k=1}(\mathbf{q}_k-\mathbf{q}_{i,l,j})(\mathbf{q}_k-\mathbf{q}_{i,l,j})^T, \mathbf{q}_{i,l,j}=\frac{1}{\kappa}\sum^{\kappa}_{k=1}\mathbf{q}_k.
\end{equation}

\noindent
The residual originated from this LiDAR camera correspondence is defined as
\begin{equation}\label{eq:lc_residual}
    \mathbf{r}_{i,l,j}=\mathbf{n}_{i,l,j}^T\left(\prescript{\mathbf{I}_{l,j}}{}{\mathbf{p}_i}-\mathbf{q}_{i,l,j}\right).
\end{equation}

\noindent
Collecting all such correspondences, the extrinsic $\mathcal{E}_C$ calibration problem could be formulated as
\begin{equation}\label{eq:lc_formulation}
    \mathcal{E}_C^\ast=\arg\min_{\mathcal{E}_C}\sum_i\sum_{\mathbf{I}_{l,j}\in\mathcal{I}_i}\left(\mathbf{n}_{i,l,j}^T\left(\prescript{\mathbf{I}_{l,j}}{}{\mathbf{p}_i}-\mathbf{q}_{i,l,j}\right)\right).
\end{equation}

Inspecting the residual in~\eqref{eq:lc_residual}, we find the $\prescript{\mathbf{I}_{l,j}}{}{\mathbf{p}_i}$ is dependent on LiDAR poses $\prescript{G}{L_0}{\mathbf{T}}_{t_j}$. This is due to the reason that LiDARs may have FoV overlap with cameras at different times (as in Fig.~\ref{fig:rotate_fov}). Since $\prescript{G}{L_0}{\mathbf{T}}_{t_j} \in \mathcal{S}$ has been well estimated from Sec.~\ref{sec:methodology_multi_lidar}, we keep them fixed in this step. Moreover, the $\mathbf{n}_{i,l,j}$ and $\mathbf{q}_{i,l,j}$ are also implicitly dependent on $\mathcal{E}_C$, since both $\mathbf{n}_{i,l,j}$ and $\mathbf{q}_{i,l,j}$ are 
related with nearest neighbor search. The complete derivative of~\eqref{eq:lc_formulation} to the variable $\mathcal{E}_C$ would be too complicated. In this paper, to simplify the optimization problem, we ignore the influence of camera extrinsic on $\mathbf{n}_{i,l,j}$ and $\mathbf{q}_{i,l,j}$. This strategy works well in practice as detailed in Sec.~\ref{sec:multi_sensor_calib}.

The non-linear optimization~\eqref{eq:lc_formulation} is solved with LM method by approximating the residuals with their first order derivatives \eqref{eq:JJJr}. The optimal $\mathcal{E}_C^\ast$ is then obtained by iteratively solving~\eqref{eq:JJJr} and updating $\delta\mathbf{x}$ to $\mathbf{x}$ using the $\boxplus$ operation~\cite{boxplus}.
\begin{equation}\label{eq:JJJr}
    \delta\mathbf{x}=-\left(\mathbf{J}^T\mathbf{J}+\mu\mathbf{I}\right)^{-1}\mathbf{J}^T\mathbf{r},
\end{equation}

\noindent
where
\begin{equation*}
\begin{aligned}
    \delta\mathbf{x}&=\begin{bmatrix}
    \cdots\ \prescript{C_l}{L_0}{\boldsymbol{\phi}}^T\ \delta\mkern-3mu\prescript{C_l}{L_0}{\mathbf{t}}^T\ \cdots
    \end{bmatrix}^T\in\mathbb{R}^{6h}\\
    \mathbf{x}&=\begin{bmatrix}
    \cdots\ \prescript{C_l}{L_0}{\mathbf{R}}\ \prescript{C_l}{L_0}{\mathbf{t}}\ \cdots
    \end{bmatrix}\\
    \mathbf{J}&=\begin{bmatrix} \cdots & \mathbf{J}_p^T & \cdots \end{bmatrix}^T,
    \mathbf{r}=\begin{bmatrix} \cdots & \mathbf{r}_p & \cdots \end{bmatrix}^T,
\end{aligned}
\end{equation*}

\noindent
with $\mathbf{J}_p$ and $\mathbf{r}_p$ being the sum of $\mathbf{J}_{i,l,j}$ and $\mathbf{r}_{i,l,j}$ when $l=p$:
\begin{equation}
\begin{aligned}
    \mathbf{J}_{i,l,j}&=\mathbf{n}_{i,l,j}^T\frac{\partial\mathbf{f(p)}}{\partial\mathbf{p}}\frac{\partial\boldsymbol{\pi(\mathbf{P})}}{\partial\mathbf{P}}
    \begin{bmatrix}
    -\prescript{C_l}{L_0}{\mathbf{R}}\left(\prescript{L_0}{}{\mathbf{p}}_i\right)^\wedge & \mathbf{I}
    \end{bmatrix}\in\mathbb{R}^{1\times6} \\
    \prescript{L_0}{}{\mathbf{p}}_i&=\left(\prescript{G}{L_0}{\mathbf{T}_{t_j}}\right)^{-1}\prescript{G}{}{\mathbf{p}_i}.
\end{aligned}
\end{equation}

\subsection{Calibration Pipeline}\label{sec:pipeline}
\begin{figure}[t]
    \centering
    \includegraphics[width=1.0\linewidth]{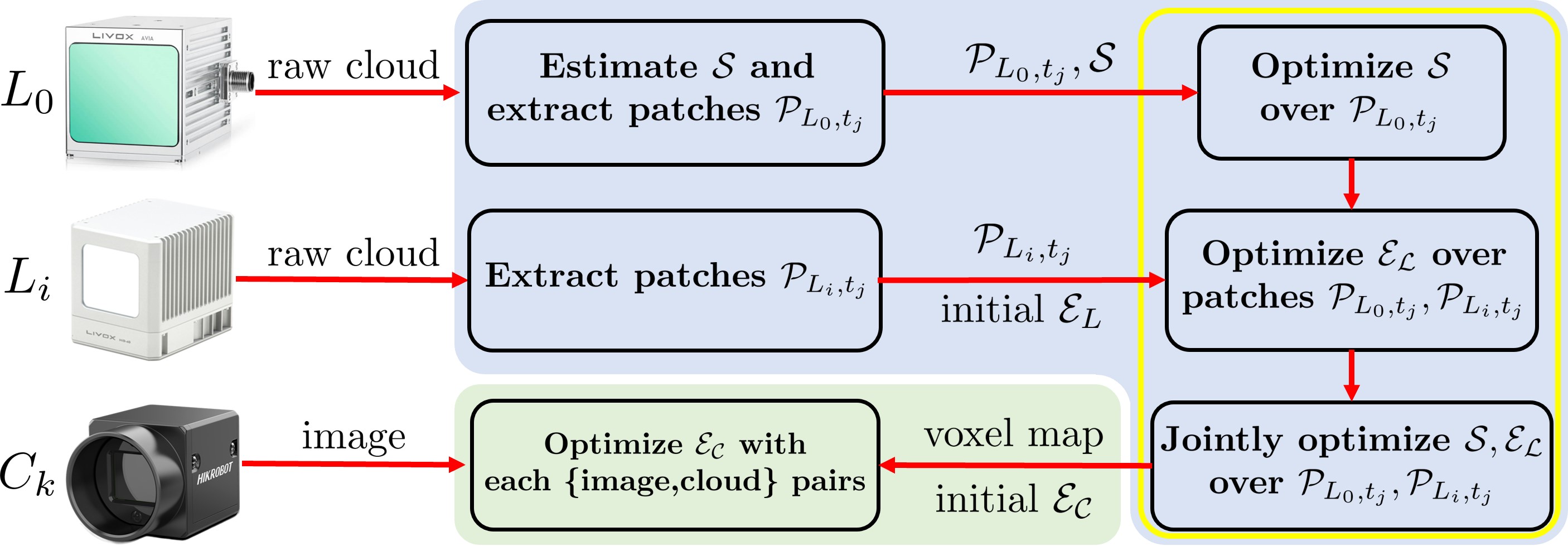}
    \caption{The workflow of our proposed method: multi-LiDAR extrinsic calibration (light blue region) and LiDAR-camera extrinsic calibration (light green region). The adaptive voxelization takes effect in the steps surrounded by the yellow rectangle.}
    \label{fig:workflow}
\end{figure}

The workflow of our proposed multi-sensor calibration is illustrated in Fig.~\ref{fig:workflow}. At the beginning of the calibration, the base LiDAR's raw point cloud is processed by a LOAM algorithm~\cite{loamlivox} to obtain the initial base LiDAR trajectory $\mathcal{S}$. Then, the raw point cloud of all LiDARs are segmented by time into point cloud patches, i.e., $\mathcal{P}_{L_i,t_j},L_i\in\mathcal{L},t_j\in\mathcal{T}$ that is collected under the pose $\prescript{G}{L_i}{\mathbf{T}}_{t_j}$.

In multi-LiDAR extrinsic calibration, the base LiDAR poses $\mathcal{S}$ are first optimized using the base LiDAR's point cloud patches $\mathcal{P}_{L_0,t_j}$. It is noticed that only $\mathcal{S}$ is involved and optimized in~\eqref{eq:lambda_3_A}. Then the extrinsic $\mathcal{E}_L$ are calibrated by aligning the point cloud from the LiDAR to be calibrated with those from the base LiDAR. In this stage's problem formulation~\eqref{eq:lambda_3_A}, $\mathcal{S}$ is fixed at the optimized values from the previous stage, and only $\mathcal{E}_L$ is optimized. Finally, both $\mathcal{S}$ and $\mathcal{E}_L$ are jointly optimized using the entire point cloud patches. In each iteration of the optimization (over $\mathcal{S}$, $\mathcal{E}_L$, or both), the adaptive voxelization (as described in Sec.~\ref{sec:voxel}) is performed with the current value of $\mathcal{S}$ and $\mathcal{E}_L$. Moreover, the Hessian matrix $\mathbf H$  has a computation complexity of $O(N^2)$, where $N$ is the number of points. In practice, to reduce this computational complexity, we down-sample the number of points scanned from the same LiDAR to 4 in each voxel. Such a process would lower the time complexity of the proposed algorithm to $O(N_{voxel})$, where $N_{voxel}$ is the total number of adaptive voxels. In Sec.~\ref{sec:multi_lidar_precision} experiment (2), $N_{voxel}\approx9\times10^3$ which is greatly smaller than the total number of raw LiDAR points in this scene, i.e., $N_{points}\approx4\times10^7$.

In multi-LiDAR-camera extrinsic calibration, the adaptive voxel map obtained with the $\mathcal{S}^\ast$ and $\mathcal{E}_L^\ast$ in the previous step is used to extract the depth-continuous edges (Sec. \ref{sec:methodology_lidar_camera}). Then those three-dimension edges are back-projected onto each image using the extrinsic parameter $\mathcal{E}_C$ and are matched with two-dimension Canny edges extracted from the image. By minimizing the residuals defined by these two edges, we iteratively solve for the optimal $\mathcal{E}_C^\ast$ with the Ceres Solver\footnote{http://ceres-solver.org/}.

\section{Experiments and Results}\label{sec:experiments}

\begin{figure}[t]{
\centering
\subfigure[]{
\begin{minipage}[t]{0.48\linewidth}
\centering
\includegraphics[width=1\linewidth]{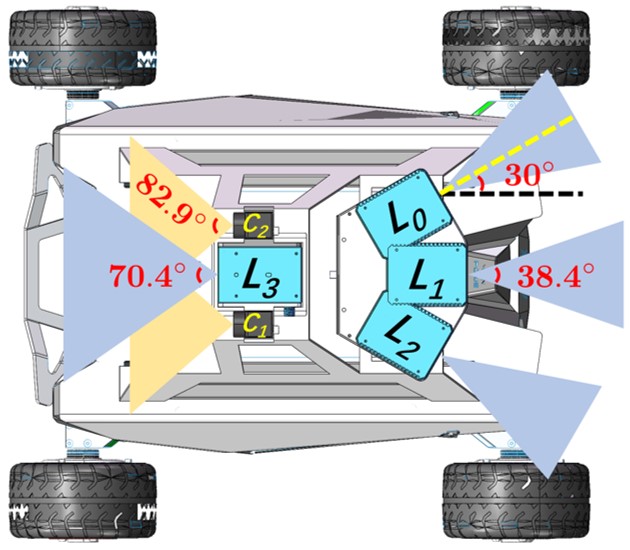}
\vspace{-0.5cm}
\label{fig:multi_fov}
\end{minipage}
}%
\subfigure[]{
\begin{minipage}[t]{0.48\linewidth}
\centering
\includegraphics[width=1\linewidth]{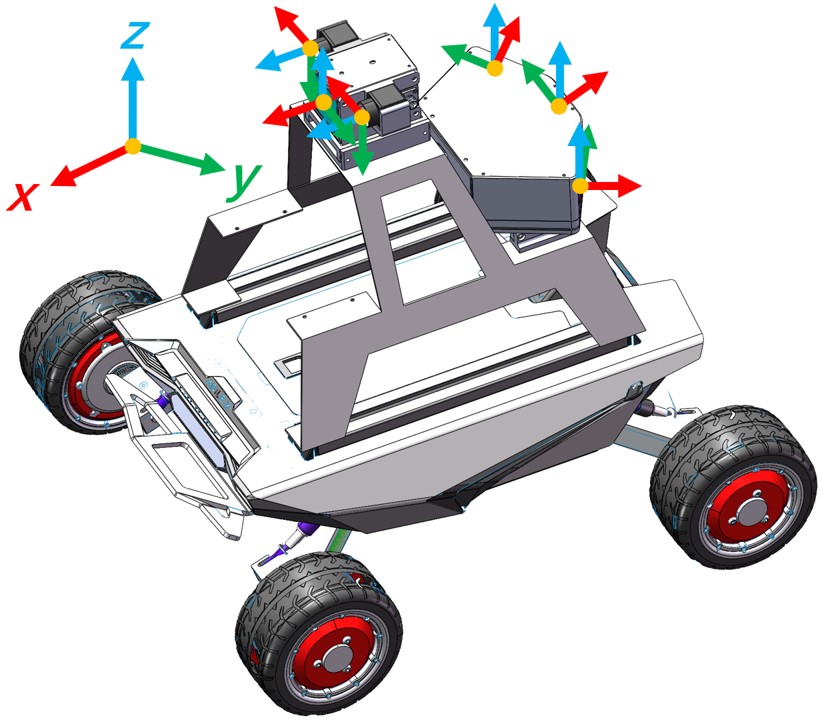}
\vspace{-0.5cm}
\label{fig:vehicle}
\end{minipage}%
}%
}
\centering
\vspace{-0.2cm}
\caption{Our customized multi-sensor vehicle platform. Left: the FoV coverage of each sensor with their FoV specs. Right: the orientation of each sensor is denoted in the right-handed coordinate system.}
\label{fig:lidar_platform}
\end{figure}

To test the proposed algorithm, we customized a remotely operated vehicle platform\footnote{https://www.agilex.ai/product/3?lang=en-us} (see Fig.~\ref{fig:lidar_platform}) with one Livox AVIA LiDAR\footnote{https://www.livoxtech.com/avia} (with 70.4 degrees of FoV, see $L_3$ in Fig.~\ref{fig:lidar_platform}), one Livox MID-100 LiDAR\footnote{https://www.livoxtech.com/mid-40-and-mid-100} (which has three internal MID-40 LiDARs, each has 38.4 degrees of FoV with only 8.4 degrees overlap between adjacent MID-40 units, see $L_0, L_1$, and $L_2$ in Fig.~\ref{fig:lidar_platform}) and two MV-CA013-21UC\footnote{https://www.rmaelectronics.com/hikrobot-mv-ca013-21uc/} cameras (with 82.9 degrees of FoV each, see $C_1$ and $C_2$ in Fig.~\ref{fig:lidar_platform}). The extrinsic parameters of the three MID-40 inside the MID-100 have been calibrated by the manufacturer and could be used as the ground truth for the calibration evaluation. 

\begin{figure}[!t]{
\centering
\subfigure[Scene-1]{
\begin{minipage}[t]{0.48\linewidth}
\centering
\includegraphics[width=1\linewidth]{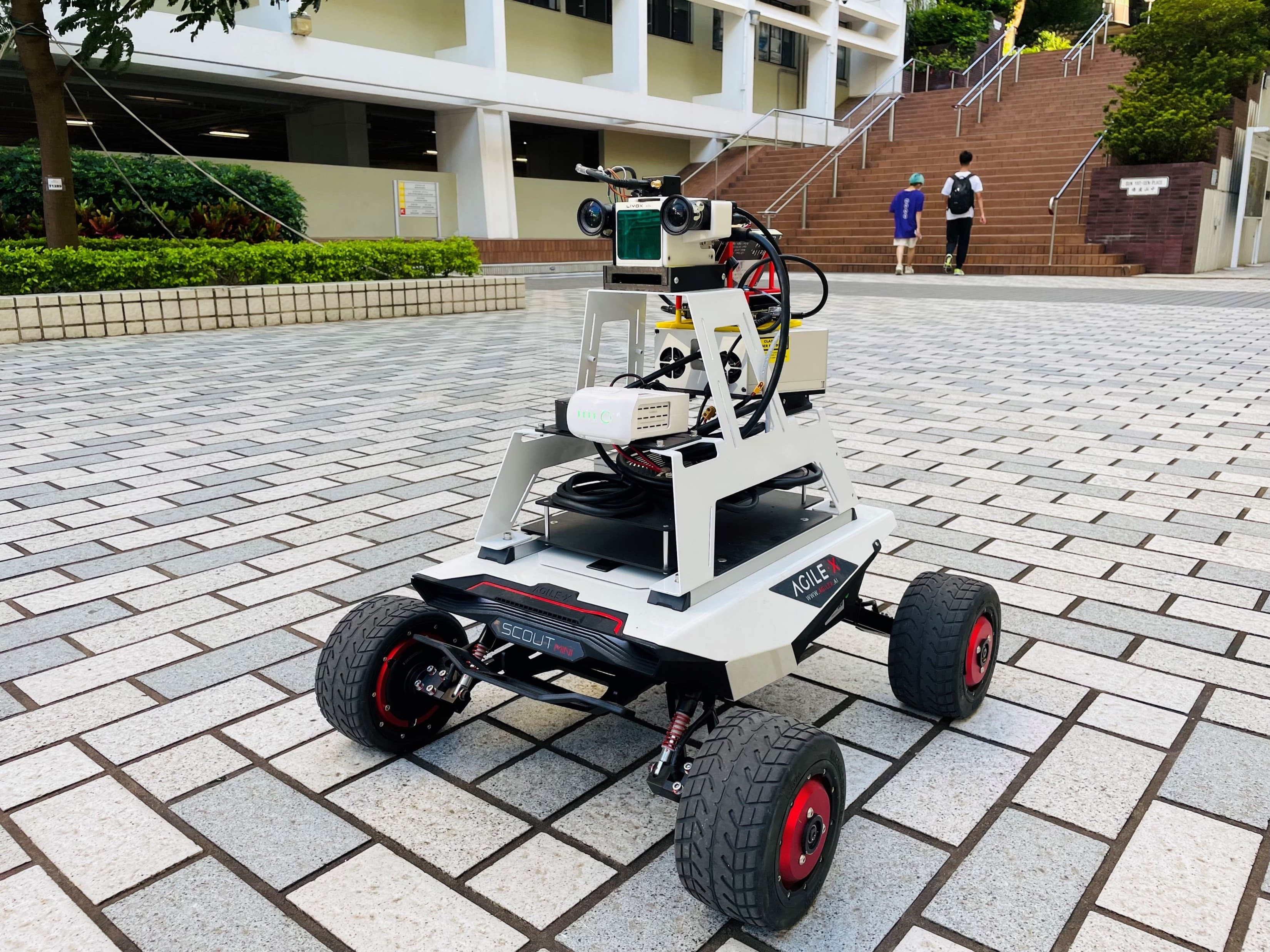}
\vspace{-0.5cm}
\label{fig:scene_1}
\end{minipage}
}%
\subfigure[Scene-2]{
\begin{minipage}[t]{0.48\linewidth}
\centering
\includegraphics[width=1\linewidth]{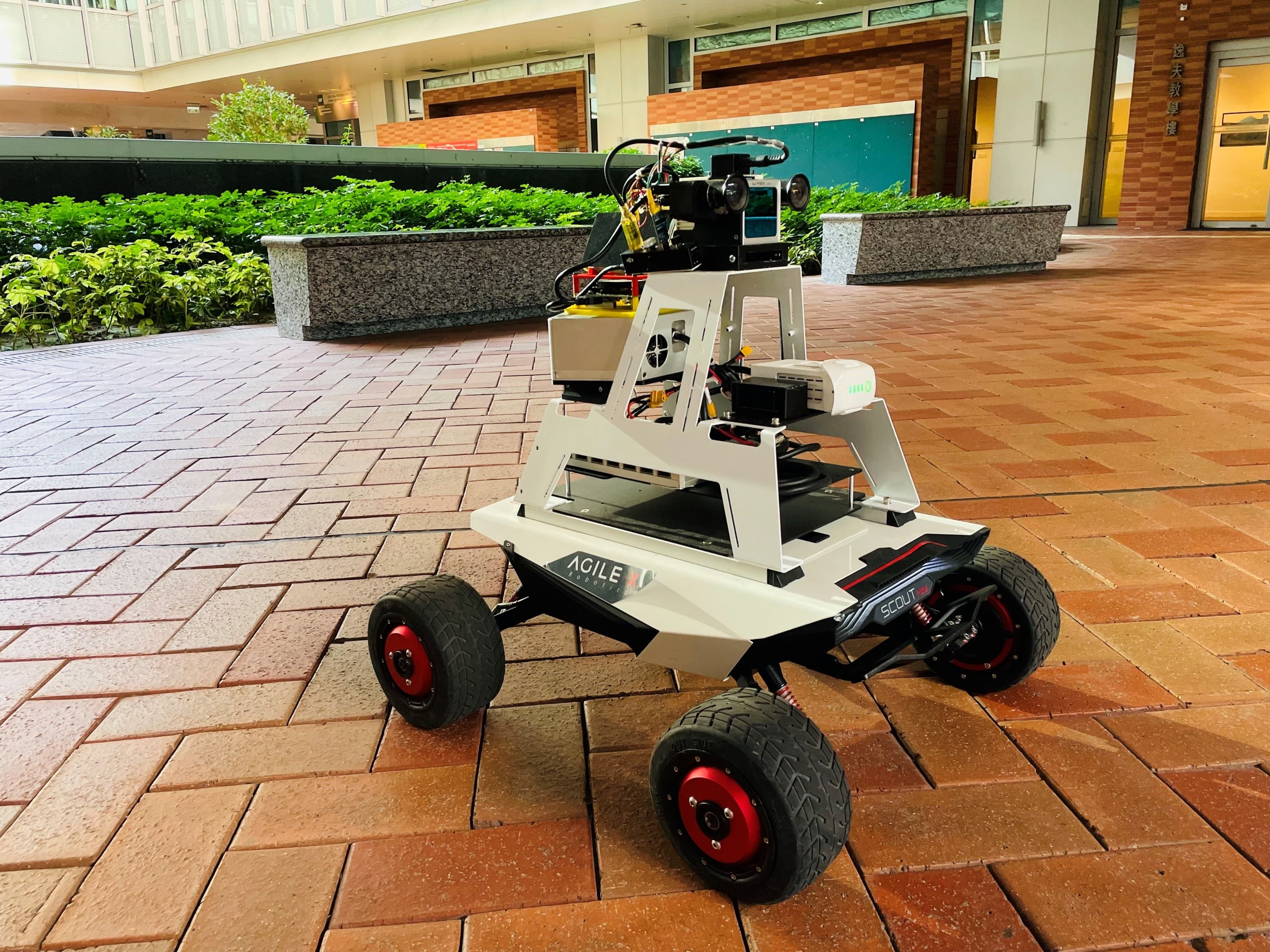}
\vspace{-0.5cm}
\label{fig:scene_2}
\end{minipage}%
}%
}
\centering
\vspace{-0.2cm}
\caption{Our experiment test scenes.}
\label{fig:two_scenes}
\end{figure}

We have verified our proposed work with the data collected in two random test scenes on our campus, as shown in Fig.~\ref{fig:two_scenes}. Scene-1 is a square in front of the main library building with moving pedestrians, and scene-2 is an open area near a garden. The calibration data is collected in both scenes by rotating the sensor suite slightly for more than 360$^\circ$ degrees and keeping this platform still every few degrees. Keeping the robot platform still during data collection enables us to acquire a dense enough point cloud from each LiDAR at each pose and also eliminates the problem caused by motion distortion and time synchronization. The timestamps $\mathcal{T}$ are manually selected so that only the point cloud and image data are chosen when the robot platform is still. During sensor suite rotation, a dedicated LiDAR inertial odometry and mapping (LOAM) algorithm loam-livox \cite{loamlivox} is called to estimate a rough LiDAR pose trajectory $\mathcal{S}_{\text{init}}$, which serves as the initial pose in our factor graph optimization. Moreover, to obtain an initial estimate of the extrinsic $\mathcal{E}_{L_{\text{init}}}$, we collected another data, the initialization data, which is collected in scene-1 with an $`$8'-figure path. Similarly, loam-livox is used to estimate each LiDAR's trajectory, based on which the extrinsic is solved by a standard Hand-eye calibration. All experiments are conducted on a desktop computer with an i7-9700K processor and 32GB RAM.

\subsection{Multiple LiDAR Calibration}\label{sec:multi_lidar_calib}
\subsubsection{Calibration Precision}\label{sec:multi_lidar_precision}
In this section, we compare our algorithm with the motion-based method~\cite{tro_motion} and the ICP-based method~\cite{previous_work} using the MID-100 LiDAR and the AVIA LiDAR. Both~\cite{tro_motion,previous_work} are targetless, offline, and utilize the motion information to calibrate the extrinsic parameter without requiring significant LiDAR FoV overlap as our method in this work. The method in \cite{tro_motion} is essentially a variant of hand-eye calibration, with further consideration of pose uncertainty. To compare the performance of \cite{tro_motion} with our method on the calibration data collected above, we run loam-livox \cite{loamlivox} to obtain the point cloud of each scan and its poses (odometry). Since the uncertainty information within each LiDAR's motion estimation is also examined in~\cite{tro_motion} to achieve the optimal performance, we manually calculate the measurement noise in each LiDAR odometry estimation. This is completed by calculating the covariance between the consecutive scans using the above-obtained point cloud and taking the odometry as the initial guess. Then, the odometry and its uncertainty information of each LiDAR are fed to and processed by~\cite{tro_motion}. The other method under comparison is our previous work~\cite{previous_work}, which used the same rotation to create an overlap for small FoV LiDARs and an ICP-based factor graph optimization to estimate the extrinsic parameter and LiDAR pose. To make a fair comparison, we feed the same initial pose trajectory $\mathcal{S}_{{\text{init}}}$ and extrinsic $\mathcal{E}_{L_{\text{init}}}$ obtained above to both \cite{previous_work} and this work before the full calibration on the calibration data collected above.

The experiment is divided into two parts: (1) MID-100 LiDAR self calibration: the middle MID-40 is chosen as the base LiDAR to calibrate the extrinsic $\mathcal{E}_L$ of the other two MID-40s, i.e., $\prescript{L_1}{L_0}{\mathbf{T}},\prescript{L_1}{L_2}{\mathbf{T}}$ (see Fig.~\ref{fig:lidar_platform}). To evaluate the calibration precision, we compare the optimized $\prescript{L_1}{L_0}{\mathbf{T}}^\ast,\prescript{L_1}{L_2}{\mathbf{T}}^\ast$ with the ground-true values obtained from the manufacturer. To further enrich the calibration data collected above, we adopt the calibration data of MID100 in two extra scenes used in~\cite{previous_work}. This leads to four test scenes in total (two scenes in this work and two scenes from \cite{previous_work}), each has two LiDAR extrinsic ground-truth (i.e., $\prescript{L_1}{L_0}{\mathbf{T}},\prescript{L_1}{L_2}{\mathbf{T}}$) for evaluation. Consequently, we have eight independent real-world calibration data for MID40 in the evaluation. (2) AVIA and MID-100 LiDAR: the AVIA LiDAR is chosen as the base LiDAR to calibrate the extrinsic $\mathcal{E}_L$ between AVIA and each MID-40s, i.e., $\prescript{L_3}{L_0}{\mathbf{T}},\prescript{L_3}{L_1}{\mathbf{T}}$ and $\prescript{L_3}{L_2}{\mathbf{T}}$ (see Fig.~\ref{fig:lidar_platform}). To evaluate the calibration precision, we calculate the $\prescript{L_1}{L_0}{\mathbf{T}^\ast}=(\prescript{L_3}{L_1}{\mathbf{T}^\ast})^{-1}\prescript{L_3}{L_0}{\mathbf{T}^\ast}$ and $\prescript{L_1}{L_2}{\mathbf{T}^\ast}=(\prescript{L_3}{L_1}{\mathbf{T}^\ast})^{-1}\prescript{L_3}{L_2}{\mathbf{T}^\ast}$ using the above results and compare them with the ground-true values obtained from the manufacturer. This experiment is conducted with calibration data collected in the two scenes from this work only since the previous work \cite{previous_work} did not have an AVIA LiDAR. The two scenes and two LiDAR extrinsic ground-truths lead to four independent real-world calibration data in the evaluation. As a result, we have twelve independent calibration data and two LiDAR types (i.e., Livox MID-40 and AVIA) with completely different scanning patterns, point densities, and FoVs.

\begin{figure}[ht]
    \centering
    \includegraphics[width=1.0\linewidth]{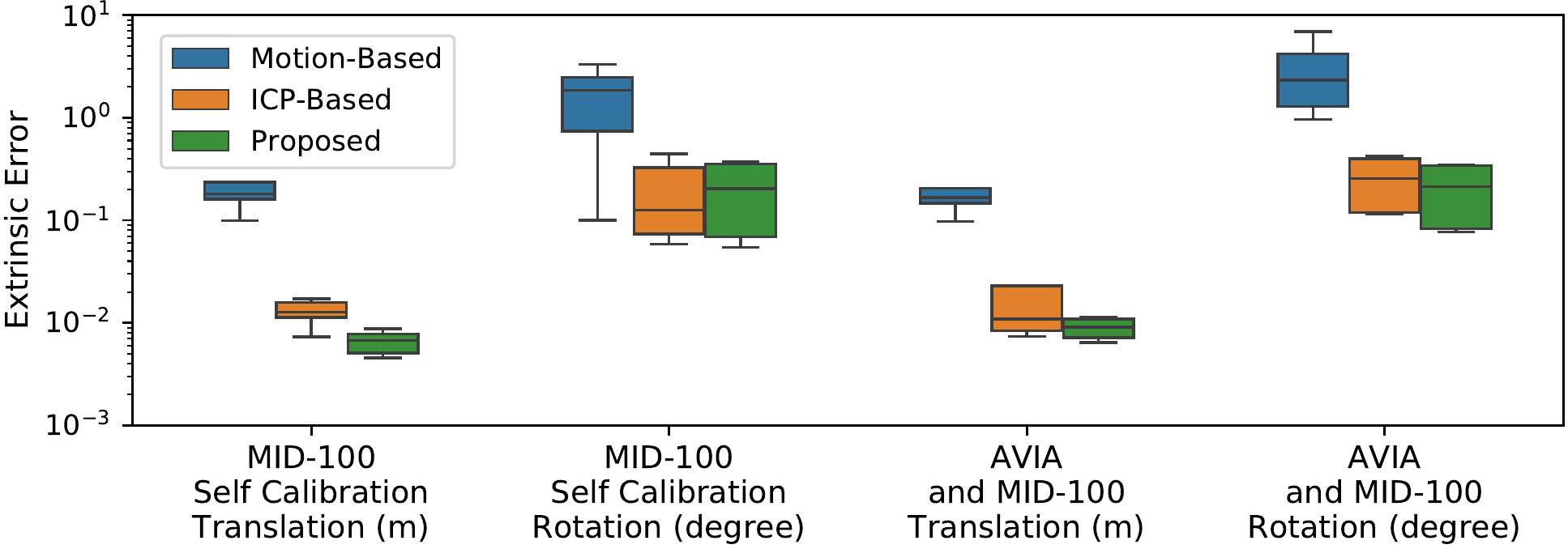}
    \caption{Extrinsic calibration results of the motion-based~\cite{tro_motion}, ICP-based~\cite{previous_work} and our proposed methods in two experiment setups (with few or no FoV overlap between sensors).}
    \label{fig:calib_precision}
\end{figure}

\begin{table*}[ht]
\centering
\caption{AVERAGE COMPUTATION TIME PER ITERATION ON MULTI-LIDAR CALIBRATION}
\label{table:compute_time}
\begin{tabular}{lccccccccc}
\toprule
\multirow{2}{*}{} & \multicolumn{3}{c}{Pose Optimization} & \multicolumn{3}{c}{Extrinsic Optimization} & \multicolumn{3}{c}{Global Optimization} \\
 & Match & Solve & Total & Match & Solve & Total & Match & Solve & Total \\
\midrule
ICP-Based~\cite{previous_work} & 4.0220s & 1.1057s & 5.1613s & 4.4635s & 1.6041s & 6.1045s & 11.0557s & 3.3616s & 14.9829s \\
\textbf{Proposed} & \textbf{0.1040s} & \textbf{0.0328s} & \textbf{0.2288s} & \textbf{0.3419s} & \textbf{0.0443s} & \textbf{0.5771s} & \textbf{2.2940s} & \textbf{0.4687s} & \textbf{3.0887s} \\
\bottomrule
\end{tabular}
\end{table*}

The comparison results of our method, the ICP-based previous work \cite{previous_work}, and motion-based \cite{tro_motion}, are shown in Fig.~\ref{fig:calib_precision}. Both the ICP-based previous method~\cite{previous_work} and our proposed method in this work outperform the motion-based method~\cite{tro_motion} in calibration precision. This is due to the reason that the precision of the motion-based method~\cite{tro_motion} relies heavily on the extent of excitation in the sensor motion. Since the sensors' movements in all test scenes are mainly constrained on the ground, as most ground vehicles do, the excitation in the z-dimension of the extrinsic is much less assured. Besides the excitation, the quality of the LiDAR pose estimation is also a crucial factor affecting the performance of hand-eye calibration in \cite{tro_motion}. For the Livox MID-40 and AVIA LiDARs that have very small FoV, the odometry is significantly deteriorated due to the reduction of feature points in a frame \cite{loamlivox}. In contrast, our current method and previous one \cite{previous_work} use a rotation motion to create a large FoV overlap, where the same feature points are observed from multiple LiDARs from multiple poses. Exploiting the constraints imposed by these co-visible features considerably increase the calibration accuracy, almost irrelevant to excitation in motion or odometry accuracy. Moreover, when compared to the ICP-based previous method \cite{previous_work}, the performance of our method in this work has considerably improved the calibration precision (in terms of average calibration error) and robustness (in terms of the variance in the calibration error), especially in translation. These results are credited to the more accurate feature matching correspondences and solutions brought by the adaptive voxelization and second-order optimization. Moreover, it is shown that our proposed method is less affected by the distinct characteristics (point cloud density, FoV, scan pattern, etc.) introduced by different types of LiDARs.

\subsubsection{Convergence and Computation Time Comparison}\label{sec:convergence_rate}
The main benefit of our method in this work, when compared to our previous work \cite{previous_work}, is the computation time, which serves as one of the main motivations for this work. In this section, we demonstrate that the proposed algorithm converges much faster than the ICP-based method~\cite{previous_work} in terms of both iteration times and computation time while remaining accurate. Since the motion-based method~\cite{tro_motion} directly generates the extrinsic result, the convergence comparison with this method is not applicable. To ensure the data diversity in the comparison, we use all the calibration data of MID-100 in the previous section collected in four scenes (two scenes collected in this work and two extra scenes from~\cite{previous_work}). We choose the middle MID-40 as the base LiDAR to calibrate the adjacent two LiDARs. To further examine the convergence robustness to initial values of the extrinsic, we perform 100 independent trials. In each trial, the initial extrinsic $\mathcal{E}_L$ is randomly perturbed ($\pm$10 degrees for $\prescript{L_1}{L_i}{\mathbf{R}}$ and $\pm$0.2m for $\prescript{L_1}{L_i}{\mathbf{t}}$) from the manufacturer's calibrated values.

\begin{figure}[t]
    \centering
    \includegraphics[width=1.0\linewidth]{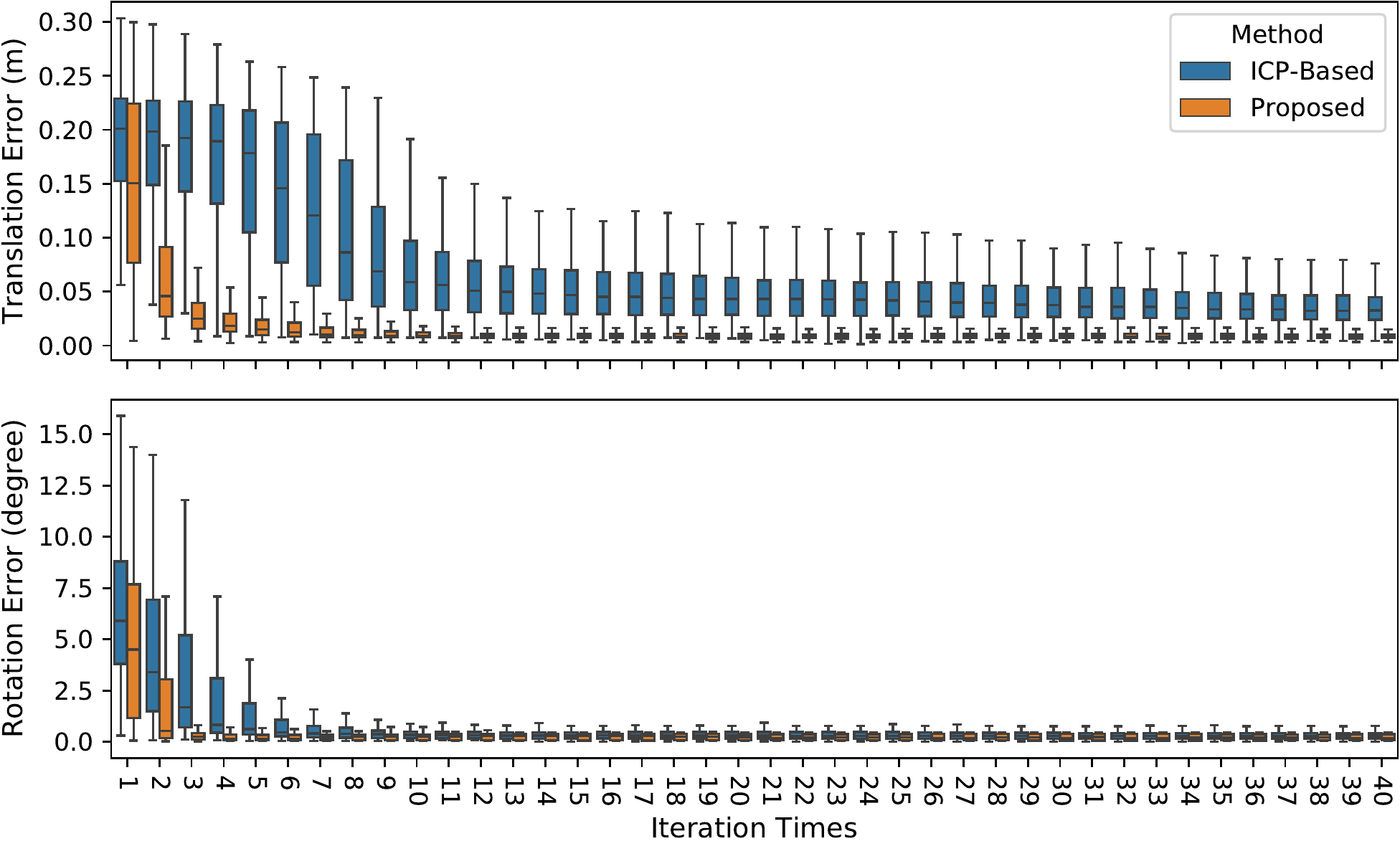}
    \caption{Convergence comparison of ICP-based~\cite{previous_work} method and our proposed work. Each box contains the results from 100 trials. The mean and standard deviation of the initial extrinsic errors are 0.1846m and 0.0562m for translation and 9.4038 degrees and 2.9094 degrees for rotation, respectively.}
    \label{fig:converge_rate}
\end{figure}

The extrinsic rotation and translation errors of both methods versus iteration numbers are plotted in Fig.~\ref{fig:converge_rate}, where the calibration error is calculated from the manufacturer values. Each box in this box-plot contains 800 calibration results from 100 trials and each trail includes the results of two LiDAR extrinsic (i.e., $\{L_0,L_1\}$ and $\{L_1,L_2\}$ see Fig.~\ref{fig:lidar_platform}) overall four scenes. As can be seen, our method converges much quicker than the previous method \cite{previous_work}, especially in translation. The entire algorithm converges within 5 iterations, even in the worst-case scenario, while that of the previous work converges much slower. After 15 iterations, the convergence in the translation of \cite{previous_work} is slowed down even more. The slow convergence of \cite{previous_work} is attributed to the pairwise ICP registration process, where only one pose or extrinsic can be estimated at a time. In contrast, our method optimizes all the poses and extrinsic concurrently, leading to a more complete point registration in each iteration and hence fewer iterations to converge. The results in Fig.~\ref{fig:converge_rate} also show how the translation error of the ICP-based method \cite{previous_work} converges to a larger value than that of our method, which is in agreement with the results in the previous section comparing the calibration prevision.

Besides the convergence rate in terms of iteration numbers, our method also achieves a much lower computation time than \cite{previous_work} at each individual iteration. The averaged computation time per trial per iteration of both methods is summarized in Table~\ref{table:compute_time}. Within each step of the calibration (see Fig.~\ref{fig:workflow}), we further dig into and calculate the time cost in feature correspondence matching (Match) and non-linear cost function solving (Solve). It is seen our proposed work significantly saves the computation time in the above two processes due to the implementation of adaptive voxelization (Sec.~\ref{sec:voxel}) and second-order optimization (Sec.~\ref{sec:methodology_multi_lidar}). In each iteration, a voxel map is created only once for our proposed work, and for any feature point, its corresponding feature points are simply the points within the same voxel. Whereas in~\cite{previous_work}, a unique $k$-d tree data structure needs to be created and searched each time for every feature point during the feature correspondence matching process. In non-linear cost function solving, the Jacobian and Hessian matrix w.r.t. the optimization variables ($\mathcal{S}$ and $\mathcal{E}_L$) are exactly derived in our proposed work, leading to a faster and more accurate solution. In contrast, in~\cite{previous_work}, only the Jacobian of the residual w.r.t. one LiDAR is considered, causing inaccurate Hessian matrix computation. This analysis is also verified in Fig.~\ref{fig:converge_rate} that the proposed work makes both the extrinsic translation and rotation errors quickly converge to the appropriate values. The reduction of iteration numbers (more than 3 times) and computation time per iteration (more than 5 times) shorten the calibration time of the previous method \cite{previous_work} by more than 15 times.

\begin{figure}[!t]
    \centering
    \includegraphics[width=1.0\linewidth]{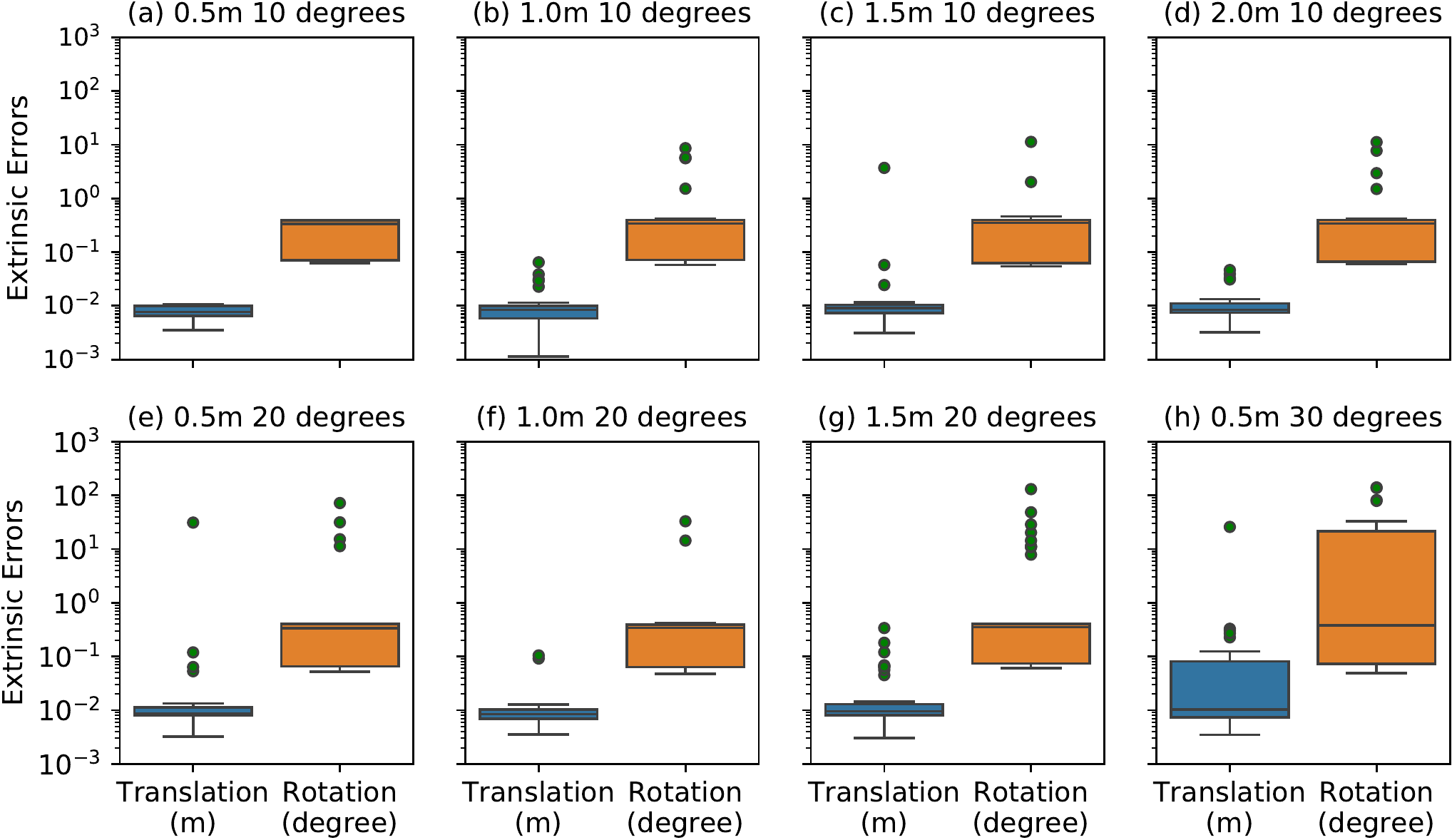}
    \caption{The distribution of calibration errors of our proposed method under multiple disturbed initial values. Each box contains the results of 100 trials. It is seen under the setups of (a)-(g), most of the initial values could be converged with only a few outliers. The initial disturbance exceeds our convergence tolerance under the setup of (h).}
    \label{fig:disturbance_test}
\end{figure}

\subsubsection{Robustness Test}
To quantify the convergence basin of our proposed method, we test our method when the initial extrinsic $\mathcal{E}_{L_{\text{init}}}$ is perturbed by noises at different levels. We use the MID-100 dataset collected from two test scenes in this work and choose the middle MID-40 as the base LiDAR to calibrate the adjacent two LiDARs. The calibration error is calculated similarly as in Sec.~\ref{sec:multi_lidar_precision}. In each configuration, the initial extrinsic is randomly perturbed 100 times (e.g., 0.5m 10 degrees means $\pm$10 degrees for $\prescript{L_1}{L_i}{\mathbf{R}}$ and $\pm$0.5m for $\prescript{L_1}{L_i}{\mathbf{t}}$) from the manufacturer's calibrated values.

The calibration errors are illustrated in Fig.~\ref{fig:disturbance_test}. Each box in this box-plot contains 400 results from 100 trials and each trail contains 4 results from two LiDAR pairs (i.e., $\{L_0,L_1\}$ and $\{L_1,L_2\}$ see Fig.~\ref{fig:lidar_platform}) in two test scenes. It is shown that given the rotation noise of 10 degrees, the proposed method could ideally converge when the translation noise is 0.5m and mostly converge when the translation noise is under 2.0m. When the rotation noise is 20 degrees, our proposed method could generally converge when the translation noise is under 1.0m. Such a high noise level is sufficient to cover the faulty scenarios in the real world caused by manufacturing mounting errors or severe vibration during usage.

\begin{figure*}[ht]
    \centering
    \includegraphics[width=1.0\linewidth]{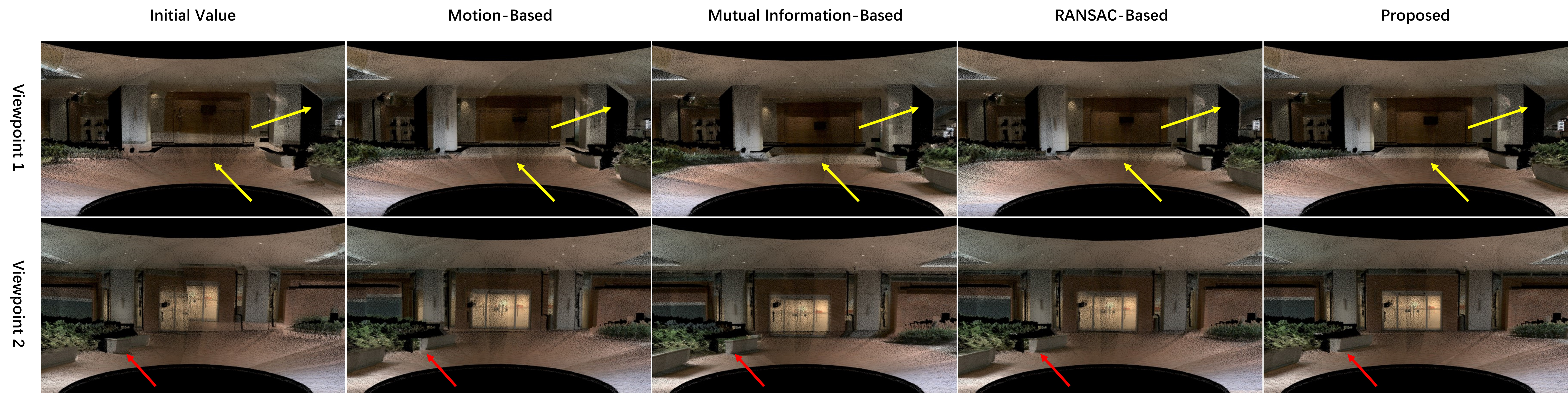}
    \caption{Point cloud colorized using the extrinsic calibrated by motion-based~\cite{tro_motion}, mutual information-based~\cite{jfr_mutual_info}, RANSAC-based~\cite{pixel_level} and our proposed methods. Each row represents a viewpoint in scene-2. The detailed difference between these methods is pointed out by arrows, e.g., miss-colorization on pillars and benches (zoomed view is recommended).}
    \label{fig:scene2_color_pc}
\end{figure*}

\begin{figure}[ht]
    \centering
    \includegraphics[width=1.0\linewidth]{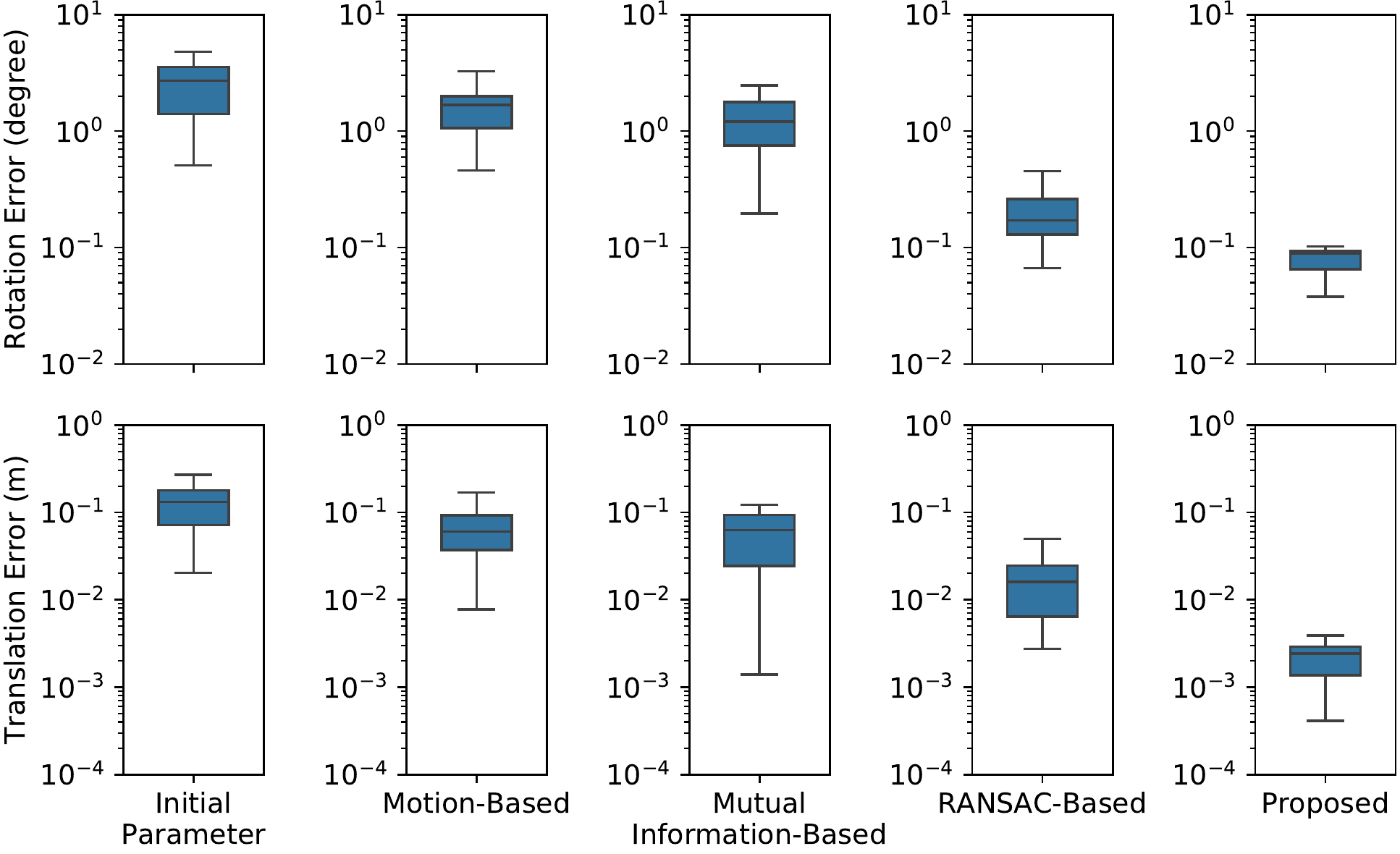}
    \caption{Extrinsic calibration results of motion-based~\cite{tro_motion}, mutual information-based~\cite{jfr_mutual_info}, RANSAC-based~\cite{pixel_level} and our proposed methods. Each box-plot illustrates the results of 50 trials using the data collected in scene-2. The mean and standard deviation of the initial rotation errors are 2.4768 and 1.2390 degrees. The mean and standard deviation of the initial translation errors are 0.1308m and 0.0682m, respectively.}
    \label{fig:avia_camera}
\end{figure}

\subsection{Multiple LiDAR Camera Calibration}\label{sec:multi_sensor_calib}
\subsubsection{LiDAR-Camera with FoV Overlap}\label{sec:with_fov_overlap}
In this section, we verify the effectiveness of our method in calibrating the extrinsic among LiDARs and cameras when they have FoV overlap. We select the AVIA as the base LiDAR and calibrate its extrinsic w.r.t. two cameras (see Fig.~\ref{fig:lidar_platform}). The extrinsic $\mathcal{E}_C$ is initialized by adding disturbance to the values measured from the CAD model. We perform 50 independent trials with the calibration data collected in scene-2, that in each trial the initial extrinsic is randomly perturbed ($\pm$5 degrees for $\prescript{C_k}{L_3}{\mathbf{R}}$ and $\pm$0.1m for $\prescript{C_k}{L_3}{\mathbf{t}}$) from the CAD model's measurements. We calibrate the extrinsic of each camera individually (i.e., $\prescript{C_1}{L_3}{\mathbf{T}^\ast},\prescript{C_2}{L_3}{\mathbf{T}^\ast}$), then we calculate the $\prescript{{C_{1}}}{{C_{2}}}{\mathbf{T}^\ast}=\prescript{C_1}{L_3}{\mathbf{T}^\ast}(\prescript{C_2}{L_3}{\mathbf{T}^\ast})^{-1}$ and compare it with that directly calibrated by the standard chessboard method serving the ground-truth.

We compare our method with three targetless methods that work for LiDAR-cameras with FoV overlaps: RANSAC-based~\cite{pixel_level}, motion-based~\cite {tro_motion}, and mutual information-based~\cite {jfr_mutual_info}. Our previous work~\cite{pixel_level} is the latest state-of-the-art specifically designed for high-resolution LiDARs, which is most similar to this work. \cite{tro_motion,jfr_mutual_info} are state-of-the-art methods originally designed for 360$^\circ$ LiDARs. In~\cite{tro_motion}, each point from a LiDAR scan is projected onto and matched with adjacent two images, and the extrinsic is optimized by minimizing the total points' color difference (i.e., the appearance) across adjacent images. 
In~\cite{jfr_mutual_info}, the extrinsic is optimized by maximizing the mutual information between LiDAR intensity images and camera images.

\begin{table*}[ht]
\centering\caption{COMPUTATION TIME ON MULTIPLE LIDAR-CAMERA CALIBRATION}\label{table:lidar_cam_time}
\begin{tabular}{lcccccc}
\toprule
& \multicolumn{3}{c}{LiDAR Feature Extraction} & \multicolumn{3}{c}{Extrinsic Optimization Per Iteration} \\
& Plane Estimation & Edge Estimation & Total & \begin{tabular}[c]{@{}c@{}}LiDAR-Camera\\ Feature Matching\end{tabular} & \begin{tabular}[c]{@{}c@{}}Solving\\ Cost Function\end{tabular} & Total \\
\midrule
Motion-Based~\cite{tro_motion} & - & - & - & 1.7690s & 3.5780s & 10.8457s \\
Mutual Information-Based~\cite{jfr_mutual_info} & - & - & - & 3.6042s & 0.6101s & 4.7520s \\
RANSAC-Based~\cite{pixel_level} & 9.6186s & 27.6738s & 37.4523s & 0.8609s & 0.5552s & 1.4548s \\
\textbf{Proposed} & \textbf{3.8054s} & \textbf{2.4494s} & \textbf{6.2892s} & \textbf{0.5424s} & \textbf{0.2510s} & \textbf{0.8278s} \\
\bottomrule
\end{tabular}
\end{table*}

The calibration results are illustrated in Fig.~\ref{fig:scene2_color_pc} and Fig.~\ref{fig:avia_camera}. It is seen that both~\cite{pixel_level} and our work are an order of magnitude better than~\cite{tro_motion,jfr_mutual_info} in both rotation and translation. This is due to the reason that the three-dimensional LiDAR edge feature is more reliable than the point cloud intensity information used in~\cite{jfr_mutual_info} and the color appearance in \cite{tro_motion}, especially in the structured test scene with large planes and long edges. This difference in calibration precision could also be visualized in Fig.~\ref{fig:scene2_color_pc}. It is also interesting to see that our work outperforms the RANSAC-based method~\cite{pixel_level} quite significantly, although they share many similarities in the overall calibration pipeline. This is due to the reason that our plane estimation method uses adaptive voxels to capture planes (and hence edges) at a finer level with higher quality than that of the fixed-size voxels used in \cite{pixel_level}. The more accurate plane and edge estimation in our method (see Fig.~\ref{fig:camera_voxel}) eventually leads to higher calibration precision and robustness.

Besides precision, our method also consumes much less computation time in each step and optimization iteration, as summarized in Table~\ref{table:lidar_cam_time}. We first compare our proposed method with~\cite{jfr_mutual_info}. Though no prior feature extraction process is needed in~\cite{jfr_mutual_info}, the calculation of the mutual information consumes significant time due to the process of all LiDAR points and image pixels. This phenomenon also appears in the motion-based method~\cite{tro_motion}, as each point from a LiDAR scan is projected onto and matched with two adjacent images. The averaged raw LiDAR points in each LiDAR scan is $N_{raw}\approx8\times10^5$ while the total number of extracted LiDAR edge feature points is $N_{feature}\approx5\times10^4$, and this discovery is in accordance with the recorded time consumption in Table~\ref{table:lidar_cam_time}.

We then compare the detailed time consumption with the RANSAC-based method~\cite{pixel_level}. In~\cite{pixel_level}, the LiDAR plane feature is extracted by first cutting the point cloud into fixed-size voxels and second analyzing the points distribution in each voxel using RANSAC. In comparison, our proposed work cuts the point cloud into voxels with sizes adapted to the environment and extracts the plane feature by analyzing eigenvalues in each voxel (see Sec.~\ref{sec:voxel}). This difference in operation also leads to distinct total voxel numbers, e.g., $N_{fixed}=9216$ versus $N_{adaptive}=1369$ in this scene, which further causes large computation time divergence in LiDAR edge feature estimation as in Table~\ref{table:lidar_cam_time}. Moreover, method in~\cite{pixel_level} are also prone to false estimations (see Fig.~\ref{fig:camera_voxel}) which makes the feature matching and cost function solving processes less reliable (see Fig.~\ref{fig:avia_camera}) and more time consuming.

\subsubsection{LiDAR-Camera without FoV Overlap}\label{sec:without_overlap}

\begin{figure}[ht]
    \centering
    \includegraphics[width=1.0\linewidth]{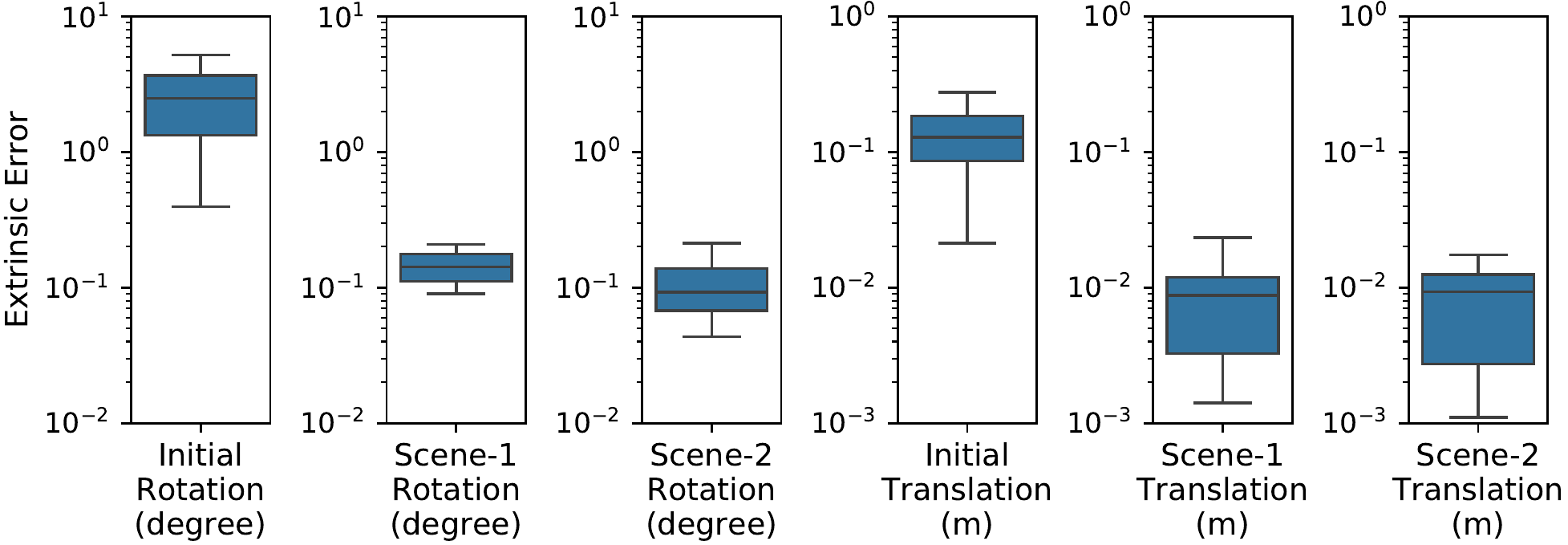}
    \caption{Extrinsic calibration results of MID-100 and opposite pointing cameras in two test scenes. Each box-plot illustrates the results of 50 trials. The mean and standard deviation of the initial rotation errors are 2.5408 and 1.3645 degrees. The mean and standard deviation of the initial translation errors are 0.1376m and 0.0663m, respectively.}
    \label{fig:mid100_camera}
\end{figure}

\begin{figure}[t]
    \centering
    \includegraphics[width=1.0\linewidth]{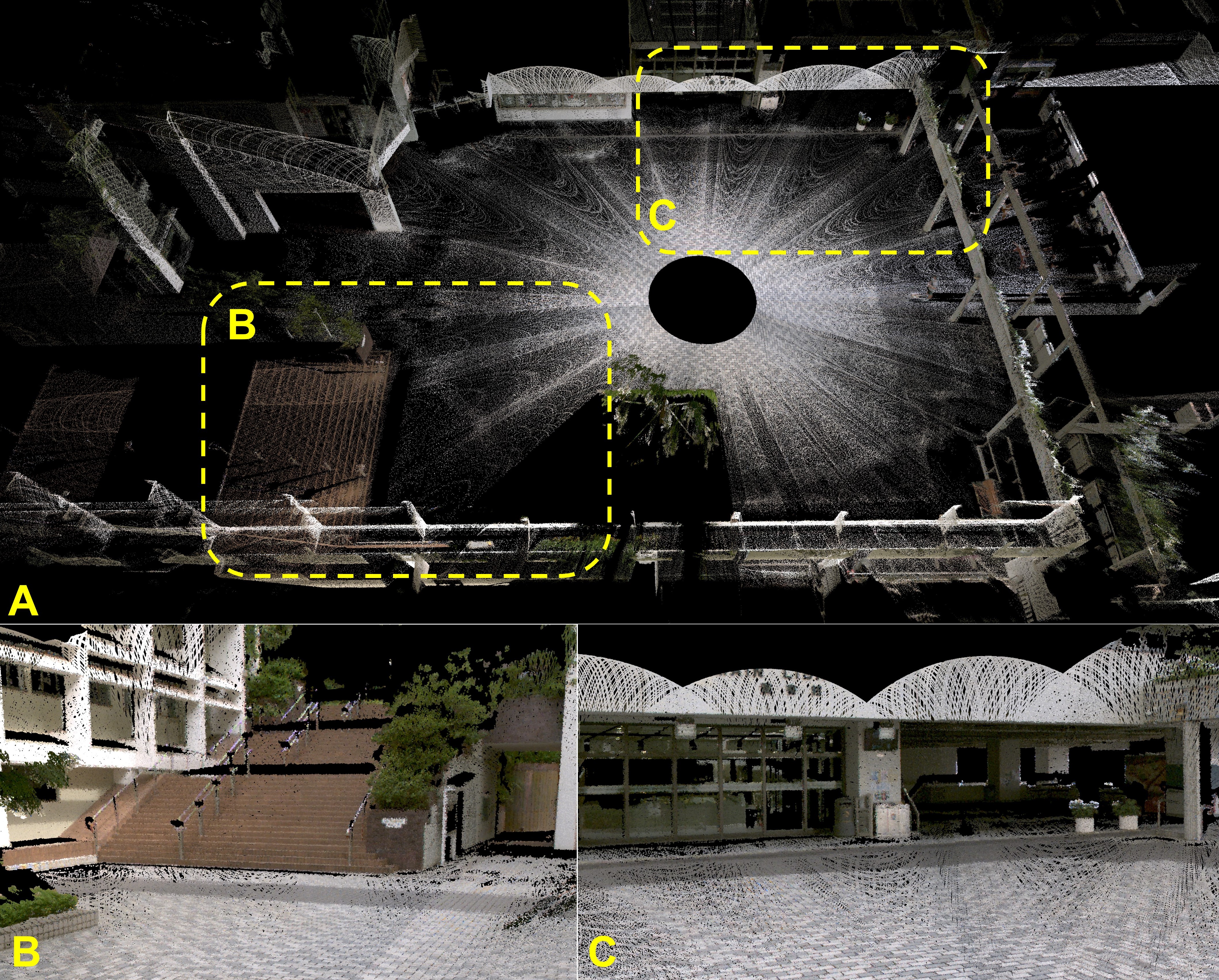}
    \caption{Colorized point cloud of MID-100 LiDAR and the opposite pointing camera in scene-1. The left camera's images are used to color the point cloud. The brightness of the building wall is due to the reflection of the sunlight. A) Bird-eye's view. B) Details of the stairs, fence, and ground tiles. C) Entrance of the library. The details of flowerpots are clearly shown.}
    \label{fig:mid100_color}
\end{figure}

In this section, we demonstrate that the proposed method could also calibrate the extrinsic $\mathcal{E}_C$ between LiDAR and cameras without FoV overlap. We choose the middle MID-40 of the MID-100 as the base LiDAR and calibrate the extrinsic of each LiDAR-camera pairs (i.e., $\prescript{C_1}{L_1}{\mathbf{T}},\prescript{C_2}{L_1}{\mathbf{T}}$, see Fig.~\ref{fig:lidar_platform}). The initial extrinsic $\mathcal{E}_C$ are calculated by adding disturbance to the values measured from the CAD model. We perform 50 independent trials with the data collected in both scenes from this work, that in each trial we randomly perturb the initial extrinsic value ($\pm$5 degrees for $\prescript{C_k}{L_1}{\mathbf{R}}$ and $\pm$0.1m for $\prescript{C_k}{L_1}{\mathbf{t}}$) from the CAD's measurements. Then we calculate the $\prescript{{C_{1}}}{{C_{2}}}{\mathbf{T}^\ast}=\prescript{C_1}{L_1}{\mathbf{T}^\ast}(\prescript{C_2}{L_1}{\mathbf{T}^\ast})^{-1}$ and compare it with that obtained by the standard chessboard method. The calibration results and the corresponding colorized point cloud are illustrated in Fig.~\ref{fig:mid100_camera} and Fig.~\ref{fig:mid100_color}.

It is seen that the general extrinsic calibration performance between MID-40 and cameras is less competitive than that between AVIA and cameras. This might be due to the reason that AVIA has larger FoV coverage (70.4 versus 38.4 degrees) and thus larger point cloud density (6 laser beams versus 1 laser beam) than MID-40, which will provide more edge correspondences in all directions. The performance of MID-40 and cameras extrinsic calibration in scene-2 is also slightly better than scene-1. This is probably due to the reason that the extracted LiDAR edges mismatch with and are trapped into the image edges that largely existed on the ground of scene-1.

\section{Conclusion}\label{sec:conclusion}
In this paper, we proposed a targetless extrinsic calibration method for multiple small FoV LiDARs and cameras. Unlike existing ICP-based methods, which rely on the $k$-d tree in LiDAR feature correspondences matching, our proposed work implemented an adaptive voxel map to store and search for the feature points to save the calibration time. We also formulated the multiple LiDAR extrinsic calibration into a Bundle Adjustment problem and derived the cost function up to second order to boost the solving process. In LiDAR-camera extrinsic calibration, we reused the above constructed adaptive voxel map to shorten LiDAR plane feature extraction and edge feature estimation time. Compared with the RANSAC-based methods, our work improved both computation efficiency and accuracy. It is believed that this open-sourced work will benefit the community of autonomous navigation robots and high-resolution mapping, especially when the sensor setups include small FoV LiDARs with few or even no FoV overlap.

Though no external calibration target is required, it is noted that our proposed work relies on the existence of natural plane features (structured building walls, ground, etc.) in the calibration environment. The precision and robustness of the extrinsic calibration among LiDARs and cameras are based on the correct extraction of LiDAR plane features. Thus, our proposed work is less reliable in unstructured scenes (e.g., country field, mountain valley, or forest). Given appropriate calibration scenes with sufficient plane features, it is believed our proposed work could produce both fast and accurate calibration results. In our future work, we wish to take the sensor measurement's noise model and camera intrinsic parameters into consideration.

\section*{Acknowledgment}
The authors gratefully acknowledge Livox Technology and AgileX Robotics for their product support. The authors would like to appreciate Zheng Liu for the insightful discussions.

\appendices
\section{}
\subsection{Elimination of Feature Parameters From Cost Function}\label{sec:app_a}
The original optimization dimension in~\eqref{eq:l_factor_app}) is too high due to the dependence on the planar parameters $\boldsymbol{\pi} = (\mathbf n_l, \mathbf q_l)$.
\begin{equation}\label{eq:l_factor_app}
    \arg\min_{\mathcal{S},\mathcal{E}_{L}, \mathbf n_l, \mathbf q_l}\sum_l \underbrace{\left(\frac{1}{N_l}\sum_{k=1}^{N_l}\left(\mathbf{n}_l^T\left(\prescript{G}{}{\mathbf{p}}_k-\mathbf{q}_l\right)\right)^2\right)}_{l\text{-th factor}}.
\end{equation}
It is noted that the planar parameters ($\mathbf n_l, \mathbf q_l$) are independent for different planes and we can optimize over them first, i.e.,
\begin{equation}\label{eq:T_n_q_app}
    \arg\min_{\mathcal{S},\mathcal{E}_{L}}\sum_l {\left(\min_{\mathbf{n}_l,\mathbf{q}_l}\frac{1}{N_l}\sum_{k=1}^{N_l}\left(\mathbf{n}_l^T\left(\prescript{G}{}{\mathbf{p}}_k-\mathbf{q}_l\right)\right)^2\right)}.
\end{equation}
The inner optimization over $(\mathbf n_l, \mathbf q_l)$ in (\ref{eq:T_n_q_app}) could be further performed on $\mathbf q_l$ first and on $\mathbf n_l$ then, i.e.,
\begin{equation}\label{seq_nq}
    \arg\min_{\mathbf{n}_l} \left( \min_{\mathbf{q}_l} \frac{1}{N_l}\sum_{k=1}^{N_l}\left(\mathbf{n}_l^T\left(\prescript{G}{}{\mathbf{p}}_k-\mathbf{q}_l\right)\right)^2  \right).
\end{equation}
As can be seen, the cost function in (\ref{seq_nq}) is quadratic w.r.t. ${\mathbf{q}_l}$. Hence the inner optimization can be solved analytically by setting the derivatives to zeros, i.e.,
\begin{equation}\label{eq:nnq0}
    \mathbf n_{l} \mathbf{n}_l^T \left( \frac{1}{N_l} \sum_{k=1}^{N_l} \left(\prescript{G}{}{\mathbf{p}}_k-\mathbf{q}_l\right) \right) = \mathbf 0.
\end{equation}
It is seen that the solution to (\ref{eq:nnq0}) is not unique as long as $\sum_{k=1}^{N_l} \left(\prescript{G}{}{\mathbf{p}}_k-\mathbf{q}_l\right)$ is perpendicular to $\mathbf n_l$, which allows $\mathbf q_l$ to move freely along any direction perpendicular to $\mathbf n_l$. Since this free movement of $\mathbf q_l$ does not change the plane parameterized by it, nor affect the cost function in (\ref{seq_nq}), any solution of $\mathbf q_l$ satisfying (\ref{eq:nnq0}) would be an optimal solution to the inner optimization problem of (\ref{seq_nq}). One such solution could be
\begin{equation}\label{eq:opt_q}
    \mathbf{q}_l^{\ast}=\frac{1}{N_l}\sum_{k=1}^{N_l}\prescript{G}{}{\mathbf{p}}_k.
\end{equation}
Substituting the optimal solution of $\mathbf{q}_l$ in (\ref{eq:opt_q}) back to (\ref{seq_nq}) leads to
\begin{equation}\label{eq:nAn}
    \arg\min_{\| \mathbf{n}_l \| = 1}\mathbf{n}_l^T\underbrace{\left(\frac{1}{N_l}\sum_{k=1}^{N_l}\prescript{G}{}{\mathbf{p}}_k\prescript{G}{}{\mathbf{p}}_k^T-\mathbf{q}_l^{\ast} \mathbf{q}_l^{\ast T}\right)}_{\mathbf{A}_l}\mathbf{n}_l.
\end{equation}
Again, this optimization problem has the well-known analytical optimal solution $\mathbf n_l^{\ast}$, which is the eigenvector corresponding to the smallest eigenvalue $\lambda_3$ of the matrix $\mathbf A_l$. As a result, substituting the optimal $\mathbf n_l^{\ast}$ back to (\ref{eq:T_n_q_app}) leads to
\begin{equation}\label{eq:lambda_3_A_app}
    \mathcal{S}^*,\mathcal{E}_L^*=\arg\min_{\mathcal{S},\mathcal{E}_{L}}\sum_l \lambda_3\left(\mathbf A_l\right).
\end{equation}
As can be seen, the optimization variables $(\mathbf n_l, \mathbf q_l)$ are analytically solved before the optimization, which significantly reduces the optimization dimension.

\subsection{Second-Order Derivation of Cost Function}\label{sec:app_b}
The optimization in (\ref{eq:lambda_3_A_app})  is nonlinear and solved iteratively. In each iteration, the cost function is approximated to the second order. More specifically, we view $\lambda_3$ as a function of all the contained points $\prescript{G}{}{\mathbf{p}}$ which is the column vector containing each $\prescript{G}{}{\mathbf{p}}_k\in\mathcal{P}_l$:
\begin{equation*}
    \prescript{G}{}{\mathbf{p}}=[\prescript{G}{}{\mathbf{p}}_1^T\prescript{G}{}{\mathbf{p}}_2^T\cdots\prescript{G}{}{\mathbf{p}}_{N_l}^T]^T \in \mathbb{R}^{3 N_l}.
\end{equation*}
The $\lambda_3(\prescript{G}{}{\mathbf{p}})$ in~\eqref{eq:lambda_3_A_app} could be approximated by
\begin{equation}\label{eq:lambda_3_Gp_app}
    \lambda_3\left(\prescript{G}{}{\mathbf{p}}+\delta\mkern-3mu\prescript{G}{}{\mkern-3mu\mathbf{p}}\right)\approx
    \lambda_3\left(\prescript{G}{}{\mathbf{p}}\right)+\mathbf{J} \cdot \delta\mkern-3mu\prescript{G}{}{\mkern-3mu\mathbf{p}}+\frac{1}{2}\mkern3mu\delta\mkern-3mu\prescript{G}{}{\mkern-3mu\mathbf{p}}^T\cdot \mathbf{H} \cdot \delta\mkern-3mu\prescript{G}{}{\mkern-3mu\mathbf{p}},
\end{equation}
where $\mathbf{J}$ and $\mathbf{H}$ are the first and second derivatives of $\lambda_3(\prescript{G}{}{\mathbf{p}})$ w.r.t. $\prescript{G}{}{\mathbf{p}}$. The expression of $\mathbf{J}$ and $\mathbf{H}$ could be found in~\cite{balm} and is omitted here due to space limit.
Suppose the $k$-th point $\prescript{G}{}{\mathbf{p}}_k$ in $\prescript{G}{}{\mathbf{p}}$ is scanned by LiDAR $L_i$ at time $t_j$, then
\begin{equation}\label{eq:rigid_transform}
\begin{aligned}
    \prescript{G}{}{\mathbf{p}}_k=&\prescript{G}{L_i}{\mathbf{T}}_{t_j} \mathbf p_k = \prescript{G}{L_0}{\mathbf{T}}_{t_j} \cdot  \prescript{L_0}{L_i}{\mathbf{T}} \cdot \mathbf p_k \\
    =& \prescript{G}{L_0}{\mathbf{R}}_{t_j}\Big(\prescript{L_0}{L_i}{\mathbf{R}}\cdot \mathbf{p}_k+\prescript{L_0}{L_i}{\mathbf{t}}\Big)+\prescript{G}{L_0}{\mathbf{t}}_{t_j},
\end{aligned}
\end{equation}
which implies $\prescript{G}{}{\mathbf{p}}_k$ is dependent on $\mathcal{S}$ and $\mathcal{E}_L$. To perturb $\prescript{G}{}{\mathbf{p}}_k$, we perturb a pose ${\mathbf{T}}$ in its tangent plane $\delta {\mathbf{T}}=[{\boldsymbol{\phi}}^T\ \delta{\mathbf{t}}^T]^T \in \mathbb{R}^6$ with the $\boxplus$ as defined in~\cite{boxplus}, i.e.,
\begin{equation}\label{eq:error_param}
\begin{aligned}
    {\mathbf{T}}&=\left({\mathbf{R}},{\mathbf{t}}\right)\\
    {\mathbf{T}}\boxplus\delta{\mathbf{T}}&=\left({\mathbf{R}}\exp\left({\boldsymbol{\phi}}^\wedge\right),{\mathbf{t}}+\delta{\mathbf{t}}\right).
\end{aligned}
\end{equation}
Based on the error parameterization in (\ref{eq:error_param}) for both $\prescript{G}{L_0}{\mathbf{T}}_{t_j}$ and extrinsic $\prescript{L_0}{L_i}{\mathbf{T}}$, the perturbed point location in (\ref{eq:rigid_transform}) is
\begin{equation}\label{eq:point_perturbed}
\begin{aligned}
    \prescript{G}{}{\mathbf{p}}_k + \delta \mkern-2mu \prescript{G}{}{\mathbf{p}}_k =&\prescript{G}{L_0}{\mathbf{R}}_{t_j}\exp\big(\prescript{G}{L_0}{\boldsymbol{\phi}}_{t_j}^\wedge\big)\Big(\prescript{L_0}{L_i}{\mathbf{R}}\exp\big(\prescript{L_0}{L_i}{\boldsymbol{\phi}}^\wedge\big)\mathbf{p}_k\\
    &+\prescript{L_0}{L_i}{\mathbf{t}}+\delta\mkern-3mu\prescript{L_0}{L_i}{\mathbf{t}}\Big)+\prescript{G}{L_0}{\mathbf{t}}_{t_j}+\delta\mkern-5mu\prescript{G}{L_0}{\mathbf{t}}_{t_j}.
\end{aligned}
\end{equation}
Then, subtracting (\ref{eq:rigid_transform}) from (\ref{eq:point_perturbed}), we obtain
\begin{equation}
\begin{aligned}
    \delta \mkern-2mu \prescript{G}{}{\mathbf{p}}_k \approx
    &\prescript{G}{L_0}{\mathbf{R}}_{t_j}\big(\prescript{L_0}{L_i}{\mathbf{R}}\mathbf{p}_k+\prescript{L_0}{L_i}{\mathbf{t}}\big)^\wedge\prescript{G}{L_0}{\boldsymbol{\phi}}_{t_j}+
    \delta\mkern-5mu\prescript{G}{L_0}{\mathbf{t}}_{t_j}+\\
    &\prescript{G}{L_i}{\mathbf{R}}_{t_j}\big(\mathbf{p}_k\big)^\wedge\prescript{L_0}{L_i}{\boldsymbol{\phi}}+ \prescript{G}{L_0}{\mathbf{R}}_{t_j}\delta\mkern-3mu\prescript{L_0}{L_i}{\mathbf{t}}
\end{aligned}
\end{equation}
and
\begin{equation}\label{eq:D}
    \begin{aligned}
        \delta\mkern-3mu\prescript{G}{}{\mkern-3mu\mathbf{p}} = \mathbf D \cdot \delta \mathbf x,
    \end{aligned}
\end{equation}
where
\begin{equation*}
\begin{aligned}
    \delta\mathbf{x}&=[\cdots\prescript{G}{L_0}{\boldsymbol{\phi}}_{t_j}^T\ \delta\mkern-5mu\prescript{G}{L_0}{\mathbf{t}}_{t_j}^T\cdots\prescript{L_0}{L_i}{\boldsymbol{\phi}}^T\ \delta\mkern-3mu\prescript{L_0}{L_i}{\mathbf{t}}^T\cdots]^T\in\mathbb{R}^{6(m+n-2)}
\end{aligned}
\end{equation*}
is a small perturbation of the optimization variable $\mathbf x$
\begin{equation*}
\begin{aligned}
    \mathbf{x}&=[\cdots\prescript{G}{L_0}{\mathbf{R}}_{t_j}\ \prescript{G}{L_0}{\mathbf{t}}_{t_j}\cdots\prescript{L_0}{L_i}{\mathbf{R}}\ \prescript{L_0}{L_i}{\mathbf{t}}\cdots], \\
\end{aligned}
\end{equation*}
and
\begin{equation}
    \begin{aligned}
    \mathbf{D}&=\begin{bmatrix}
    &\mkern-10mu\vdots&&\mkern-10mu\vdots&\\
    \cdots&\mkern-10mu\mathbf{D}_{k,p}^\mathcal{S}&\mkern-10mu\cdots&\mkern-10mu\mathbf{D}_{k,q}^{\mathcal{E}_L}&\mkern-10mu\cdots\\
    &\mkern-10mu\vdots&&\mkern-10mu\vdots&
    \end{bmatrix}\in\mathbb{R}^{3N_l\times6(m+n-2)}\\
        \mathbf{D}_{k,p}^\mathcal{S}&=\left\{
        \begin{array}{cc}
        \Big[-\prescript{G}{L_0}{\mathbf{R}}_{t_j}\big(\prescript{L_0}{L_i}{\mathbf{R}}\mathbf{p}_k+\prescript{L_0}{L_i}{\mathbf{t}}\big)^\wedge\ \mathbf{I}\Big], & \text{if }p=j\\
        \mathbf{0}_{3\times6}, & \text{else}
        \end{array}
    \right.\\
    \mathbf{D}_{k,q}^{\mathcal{E}_L}&=\left\{
    \begin{array}{cc}
        \Big[-\prescript{G}{L_0}{\mathbf{R}}_{t_j}\prescript{L_0}{L_i}{\mathbf{R}}\big(\mathbf{p}_k\big)^\wedge\ \prescript{G}{L_0}{\mathbf{R}}_{t_j}\Big], & \text{if }q=i \\
        \mathbf{0}_{3\times6}, & \text{else}.
    \end{array}
    \right.
    \end{aligned}
\end{equation}
Substituting~\eqref{eq:D} to~\eqref{eq:lambda_3_Gp_app} leads to
\begin{equation}\label{eq:lamdba_3_x_app}
\begin{split}
    \lambda_3(\mathbf{x}\boxplus\mathbf{\delta x})&\approx
    \lambda_3(\mathbf{x})+\mathbf{JD}\delta\mathbf{x}+\frac{1}{2}\delta\mathbf{x}^T\mathbf{D}^T\mathbf{HD}\delta\mathbf{x} \\
    &=\lambda_3(\mathbf{x})+\mathbf{\Bar{J}}\delta\mathbf{x}+\frac{1}{2}\delta\mathbf{x}^T\mathbf{\Bar{H}}\delta\mathbf{x}.
\end{split}
\end{equation}

\bibliographystyle{unsrt}
\bibliography{reference}

\begin{IEEEbiography}
[{\includegraphics[width=1in,height=1.25in,clip,keepaspectratio]{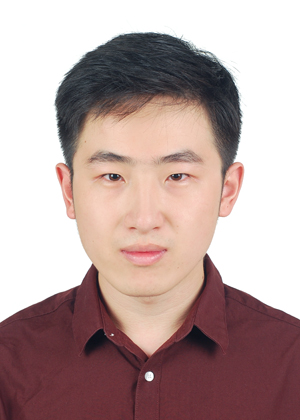}}]{Xiyuan Liu}
received the B.Eng. degree in electronic and computer engineering from Hong Kong University of Science and Technology, Hong Kong, in 2017, and the M.Phil. degree in electronic and computer engineering from Hong Kong University of Science and Technology, Hong Kong, in 2019.

He is currently a Ph.D. student with the University of Hong Kong, Hong Kong, and his research interests include LiDAR mapping and sensor calibration.
\end{IEEEbiography}

\begin{IEEEbiography}
[{\includegraphics[width=1in,height=1.25in,clip,keepaspectratio]{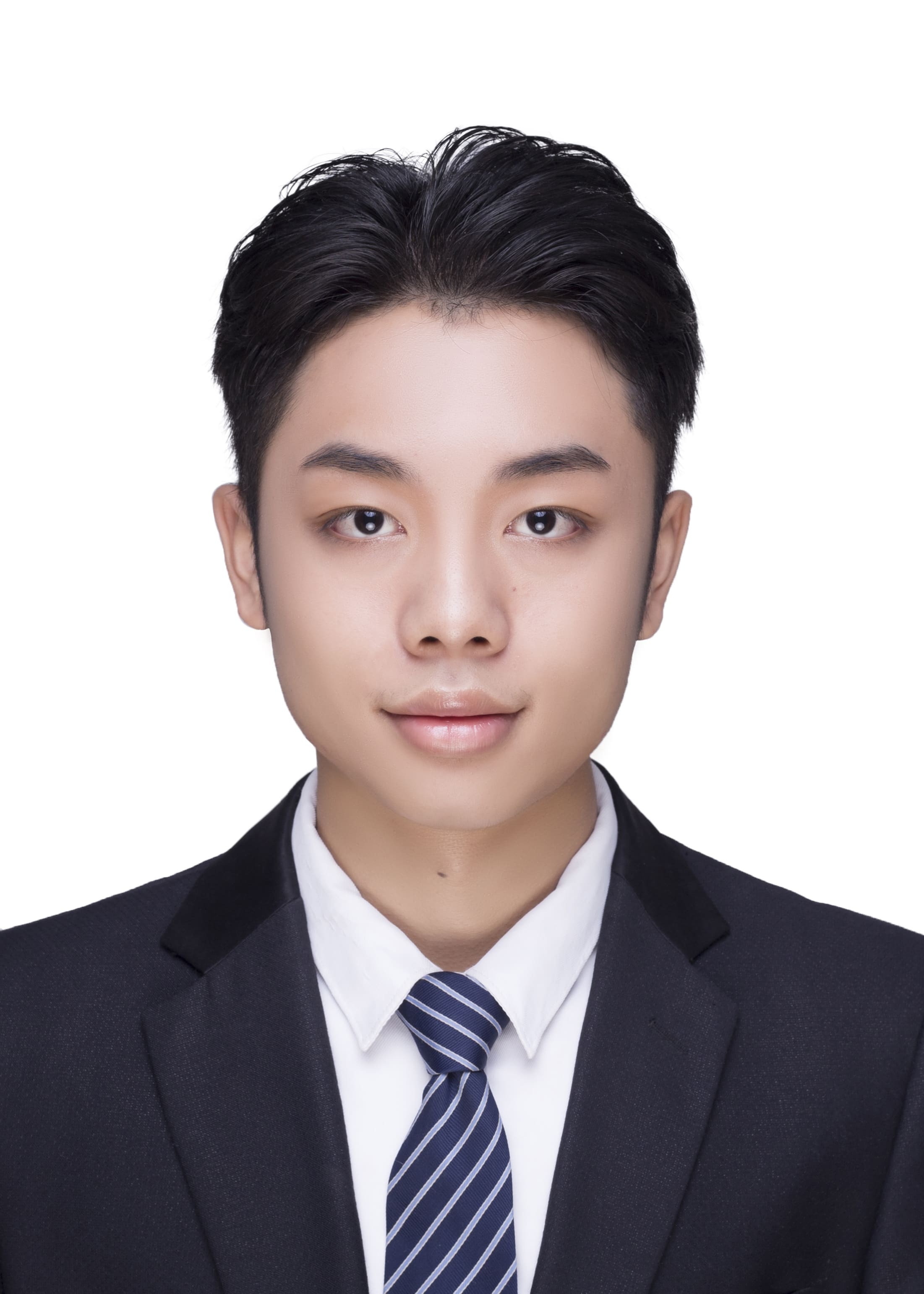}}]{Chongjian Yuan}
received the B.Eng. degree in automation from the Zhejiang University (ZJU), Hangzhou, Zhejiang, China, in 2020.

He is currently a Ph.D. student with the University of Hong Kong, Hong Kong, and his research interests include LiDAR SLAM and sensor calibration. 
\end{IEEEbiography}

\begin{IEEEbiography}
[{\includegraphics[width=1in,height=1.25in,clip,keepaspectratio]{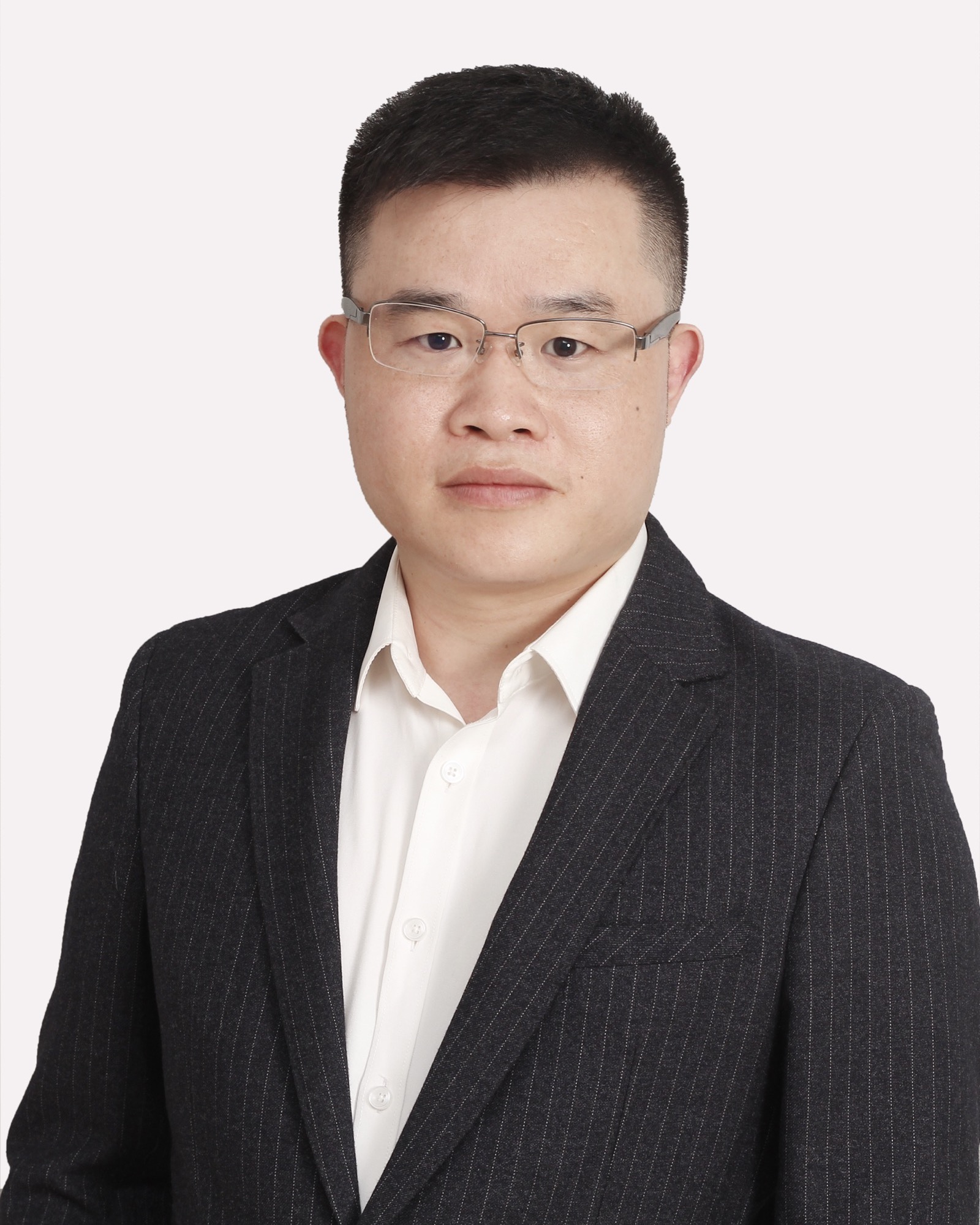}}]{Fu Zhang}
received the B.E. degree in automation from the University of Science and Technology of China (USTC), Hefei, Anhui, China, in 2011, and the Ph.D. degree in Controls from the University of California, Berkeley, CA, USA, in 2015.

He joined the department of mechanical engineering, the University of Hong Kong (HKU), as an Assistant Professor from Aug 2018. His current research interests are on robotics and controls, with focus on UAV design, navigation, control, and LiDAR-based simultaneous localization and mapping.
\end{IEEEbiography}

\end{document}